\documentclass[journal,anonymous=true]{IEEEtran}
\IEEEoverridecommandlockouts

\usepackage{cite}

\usepackage{mathrsfs}
\usepackage{lipsum}
\usepackage{graphicx}
\usepackage{epsfig}
\usepackage{float}
\usepackage{amsfonts}
\usepackage{dsfont}
\usepackage{subcaption}
\usepackage{setspace}
\usepackage[ruled,linesnumbered,noend]{algorithm2e}
\usepackage{algpseudocode}
\usepackage{threeparttable}

\usepackage{array}

\usepackage[utf8]{inputenc} 
\usepackage[T1]{fontenc}    
\usepackage{graphicx} \graphicspath{ {figures/} }
\usepackage{times}
\usepackage{graphicx}
\usepackage{amsmath,amssymb}
\usepackage{acronym}
\usepackage{enumitem}
\usepackage[breaklinks,colorlinks,linkcolor=black]{hyperref}
\usepackage{balance}
\usepackage{xspace}
\usepackage{setspace}
\usepackage[skip=3pt,font=small]{subcaption}
\usepackage[skip=3pt,font=small]{caption}
\usepackage[dvipsnames,svgnames,x11names]{xcolor}
\usepackage[capitalise,nameinlink]{cleveref}
\usepackage{booktabs,tabularx,colortbl,multirow,multicol,array,makecell}
\usepackage{dblfloatfix}
\usepackage[misc]{ifsym}
\usepackage{pifont}
\usepackage{cite}

\makeatletter
\DeclareRobustCommand\onedot{\futurelet\@let@token\@onedot}
\def\@onedot{\ifx\@let@token.\else.\null\fi\xspace}

\makeatother

\makeatletter
\def\BState{\State\hskip-\ALG@thistlm}
\makeatother

\makeatletter
\renewcommand{\paragraph}{%
  \@startsection{paragraph}{4}%
  {\z@}{0ex \@plus 0ex \@minus 0ex}{-1em}%
  {\hskip\parindent\normalfont\normalsize\bfseries}%
}
\makeatother

\crefname{algorithm}{Alg.}{Algs.}
\Crefname{algocf}{Algorithm}{Algorithms}
\crefname{section}{Sec.}{Secs.}
\Crefname{section}{Section}{Sections}
\crefname{table}{Tab.}{Tabs.}
\Crefname{table}{Table}{Tables}
\crefname{figure}{Fig.}{Fig.}

\definecolor{gblue}{HTML}{4285F4}
\definecolor{gred}{HTML}{DB4437}
\definecolor{ggreen}{HTML}{0F9D58}

\definecolor{mygray}{gray}{.92}

\begin{document}

\title{AINav: Interactive Navigation via LLM-based Primitive Skill Tree and Adaptive Replanning}

\author{Kangjie Zhou, Yao Mu, Haoyang Song, Yi Zeng, Pengying Wu, Han Gao, and Chang Liu
\thanks{This work is sponsored by Beijing Nova Program (20220484056, 20240484498), National Natural Science Foundation of China (62203018), and State Key Laboratory of Intelligent Green Vehicle and Mobility (KFZ2410) (\textit{Corresponding author: Chang Liu.})}
\thanks{Kangjie Zhou, Haoyang Song, Yi Zeng, Pengying Wu, Han Gao, and Chang Liu are with School of Advanced Manufacturing and Robotics, Peking University.}
\thanks{Yao Mu is with AI Institute, School of Computer Science, Shanghai Jiao Tong University.}
}
\maketitle

\begin{abstract}

Robotic navigation in complex environments remains a critical research challenge.
Traditional navigation methods focus on optimal trajectory generation within fixed free workspace, therefore struggling in environments lacking viable paths to the goal, such as disaster zones or cluttered warehouses. 
To address this problem, we propose AINav, an adaptive interactive navigation approach that proactively interacts with environments to create feasible paths to achieve originally unreachable goals.
Specifically, we present a primitive skill tree for task planning with large language models (LLMs), facilitating effective reasoning to determine interaction objects and sequences.
To ensure robust subtask execution, we adopt reinforcement learning to pre-train a comprehensive skill library containing versatile locomotion and interaction behaviors for motion planning.
Furthermore, we introduce an adaptive replanning approach featuring two LLM-based modules: an advisor serving as a flexible replanning trigger and an arborist for autonomous plan adjustment.
Integrated with the tree structure, the replanning mechanism allows for convenient node addition and pruning, enabling rapid plan adaptation in a priori unknown environments.
Comprehensive simulations and experiments have demonstrated AINav's effectiveness and adaptivity in diverse scenarios.
The supplementary video is available at: https://youtu.be/CjXm5KFx9AI.

\end{abstract}


\section{Introduction}

Navigation in diverse and complex environments is a pivotal problem in robotics, necessitating innovative approaches to ensure effective and adaptable decision-making and planning. 
Recent research has demonstrated the potential of quadruped robots to traverse cluttered environments and navigate to the desired position on complex terrains~\cite{hoeller2024anymal,han2024lifelike}. 
Despite the progresses, conventional navigation strategies primarily focus on optimal trajectory planning within a given free workspace, which significantly limits their applicability in real-world scenarios such as disaster rescue or warehouse logistics, where feasible paths may not exist a priori, and the robot needs to interact with the environment to create viable routes.

To tackle this challenge, previous works have explored strategies for a special case, namely navigation among movable obstacles (NAMO), where the robot can actively interact with movable obstacles to create viable paths.
However, prior research~\cite{muguira2023visibility} primarily focused on global planning and assumed deterministic object dynamics during interaction in simplified simulations, which limited their applicability in real-world situations.
To address this limitation, Simon et al.~\cite{armleder2024tactile} have explored integrating mobile platforms with robotic arms, employing whole-body control to relocate obstacles and create pathways, and modeling more realistic interaction behaviors to enhance applicability in real-world environments. 
However, NAMO approaches only focus on removing obstacles without utilization of movable obstacles as tools, such as leveraging shorter obstacles as stepping stones to help the robot traverse originally impassable hurdles, thereby limiting their abilities to solve long-horizon interactive navigation tasks that require multi-step strategic utilization of environmental objects.

\begin{figure}[!t] 
\centering
\includegraphics[width=\linewidth]{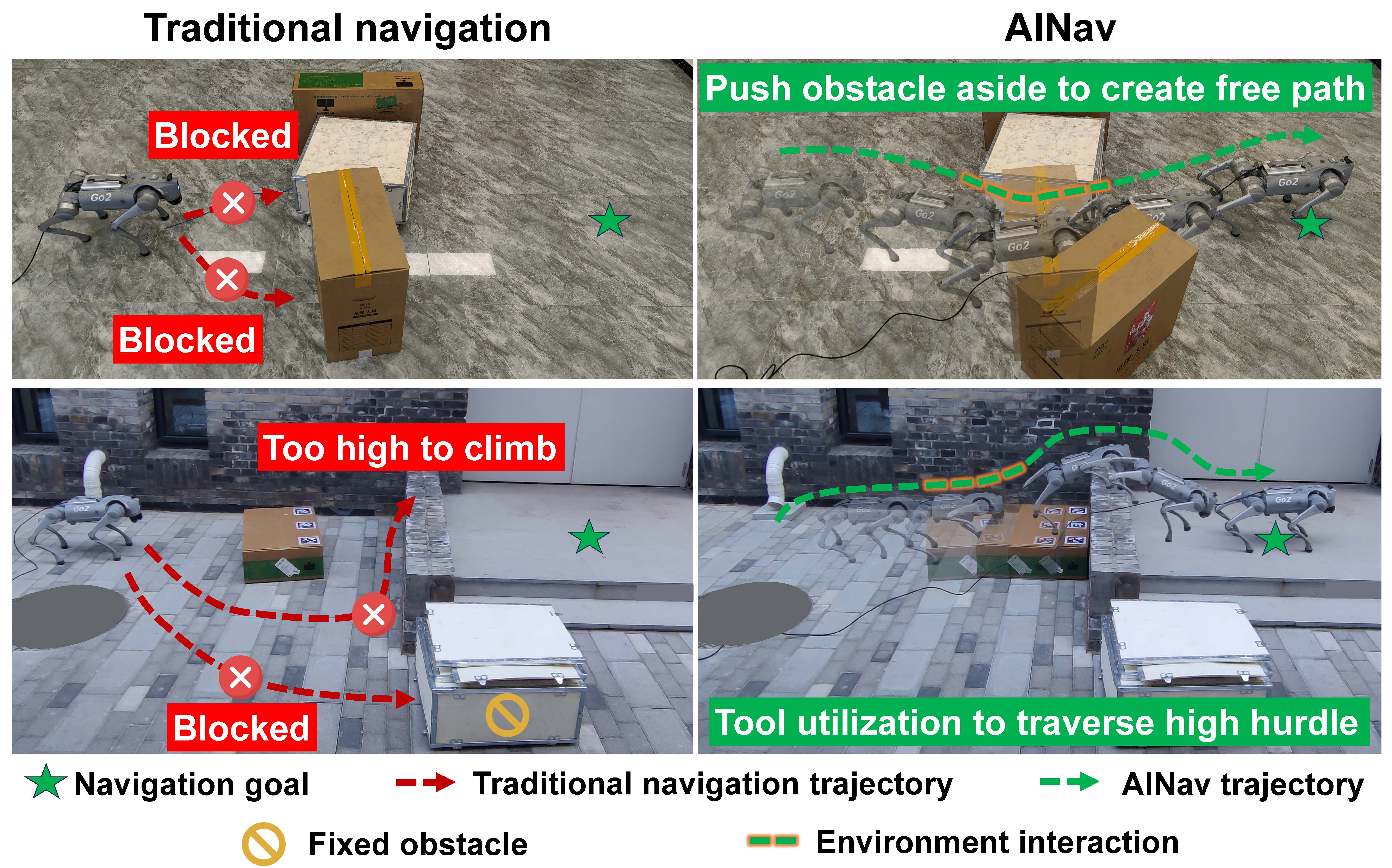} 
\caption{\textbf{Interactive navigation in challenging environments.}
In contrast to conventional obstacle-avoidance navigation systems that passively adapt to predefined free space and fail to reach goals in environments without feasible paths, AINav can push aside obstacles to create pathways within congested obstacles or utilize obstacles as tools to traverse excessively high hurdles, enabling interactive navigation in challenging environments.} 
\label{Fig: cover_pic} 
\end{figure}

Leveraging the advanced understanding and reasoning abilities of large language models (LLMs), recent works have combined LLMs with pre-trained atomic skill libraries to interact with environments to address complex long-horizon tasks~\cite{ouyang2024long,xu2023creative}. 
Specifically, the interactive tasks are formulated as task and motion planning (TAMP) problems, where task planning involves task decomposition to assist in completing final tasks, such as selecting which objects to interact with and determining the sequence of interactions, while motion planning designs controllers to execute the high-level tasks.
Existing TAMP methods usually require global scene description from an omniscient perspective, which is often unavailable in real-world scenarios where robots operate with limited, egocentric observations. 
This limitation restricts their ability to adapt to dynamic and partially observable environments.
Closed-loop task planning can respond to new observations through iterative replanning~\cite{hu2023tree,wang2023describe}.
Nevertheless, previous works commonly focus on passive failure recovery, lacking the ability to proactively utilize new information and adjust plans, which represents a critical aspect for effective navigation in partially observable environments.
Therefore, the primary challenge we aim to address is how to create effective plans for navigation tasks in unknown interactive environments while adaptively responding to newly acquired information.

To overcome these challenges, we propose the \textbf{A}daptive \textbf{I}nteractive \textbf{Nav}igation (AINav), a closed-loop system that demonstrates dual capabilities: it actively interacts with the environment to change the workspace for achieving navigation objectives, and adaptively reacts to new observations (\Cref{Fig: cover_pic}). 
This dual functionality enables effective navigation in unfamiliar and interactive environments using egocentric sensing information, without relying on global environmental knowledge.
While recent advancements in robot parkour~\cite{cheng2024extreme} have demonstrated the capability to traverse across varying heights relative to the size of the quadruped robot, our research emphasizes enabling the robot to actively manipulate the environment, tackling challenges that extend beyond the scope of conventional passive obstacle avoidance and terrain adaptation navigation systems.
The main contributions can be summarized as follows:
\begin{itemize}
\item We propose a primitive skill tree for task planning using LLM, which represents task decomposition as a tree structure with atomic skills.
This design enables robust reasoning by exploration of multiple potential solutions and rapid adaptation to new information by tree structure modification.
\item We develop a comprehensive skill library with reinforcement learning (RL) to equip the robot with advanced locomotion and interaction capabilities.
Integrated with customized reward design and task curriculum, robots can learn to navigate and interact robustly in complex environments.
\item We introduce an adaptive replanning mechanism including two LLM-based agents, an advisor and an arborist, where the advisor adaptively triggers replanning based on new observations, and the arborist automatically adjusts the decision tree for optimal task execution.
\end{itemize}

We validate AINav across various simulations and real-world experiments, revealing its efficacy in tackling challenging navigation tasks and its capacity for fast adaptation to new observations in unknown environments.
Through this novel framework, we aim to extend the boundaries of robotic navigation, offering an effective and computationally efficient solution for interactive navigation in complex environments.
The remainder of this paper is organized as follows: 
Section II reviews related work in robotic navigation and interactive planning. 
Section III introduces the problem formulation of interactive navigation.
Section IV presents the proposed AINav framework. 
Section V describes the experimental setup and results. 
Finally, Section VI concludes the manuscript and discusses future directions.

\section{Related Work}

\subsection{Robot Navigation in Complex Environments}

Navigation in complex environments requires robots to traverse cluttered spaces, dynamically avoid obstacles, and adapt to irregular terrains~\cite{hoeller2024anymal}.
Leveraging the superior agility and adaptivity of quadruped robots, significant progress has been made to complete navigation tasks across diverse complex terrain, including sampling-based methods~\cite {wellhausen2023artplanner}, learning-based methods~\cite {rudin2022advanced}.
Recently, benefiting from the powerful understanding and reasoning ability of pre-trained large models, abundant works utilized the LLM and vision-language models (VLM) to understand the semantic environment information for more flexible terrain traverse, so as to tackle navigation tasks~\cite{zhu2024cross,shah2023vint}.
However, current approaches often assume the existence of navigation paths, neglecting the possibility that such paths may not exist in real cluttered scenarios.
Zhang et al.~\cite{zhang2024interactive} presented an interactive navigation framework that instructs robots to navigate in an environment with traversable obstacles with LLMs.
However, they aim to adapt to the environment passively rather than actively utilizing interactive objects for task completion.
Wu et al.~\cite{ouyang2024long} and Xu et al.~\cite{xu2023creative} proposed using LLMs to interact with environments and employ tools to address complex long-horizon tasks with quadruped robots more effectively.
However, they assume complete scene descriptions with omniscient information and perform open-loop planning, which is impractical for navigation tasks where the robot obtains egocentric observation incrementally.

\subsection{Navigation Among Movable Obstacles}
Navigation among movable obstacles refers to the problem of robot navigation that requires manipulation of the environment.
Early research~\cite{muguira2023visibility} mainly concentrated on modeling abstract and idealized interactive behaviors, overlooking the complexities inherent in real-world physical interactions. 
While these approaches are effective in simplified simulation settings, their failure to account for real-world physics significantly restricts their practical applicability. 
Recent work by Simon et al.~\cite{armleder2024tactile} has advanced the field by integrating mobile platforms with robotic arms and utilizing whole-body control to relocate obstacles and clear pathways. 
This approach incorporates more realistic interaction models, improving applicability in real-world environments. 
Nevertheless, current NAMO approaches predominantly focus on obstacle removal and do not fully capitalize on the potential of movable obstacles, such as utilizing shorter obstacles as steps to traverse higher barriers. 
This limitation restricts their ability to address more complex long-horizon tasks in more intricate scenarios.


\begin{figure*}[!t] 
\centering
\includegraphics[width=0.97\linewidth]{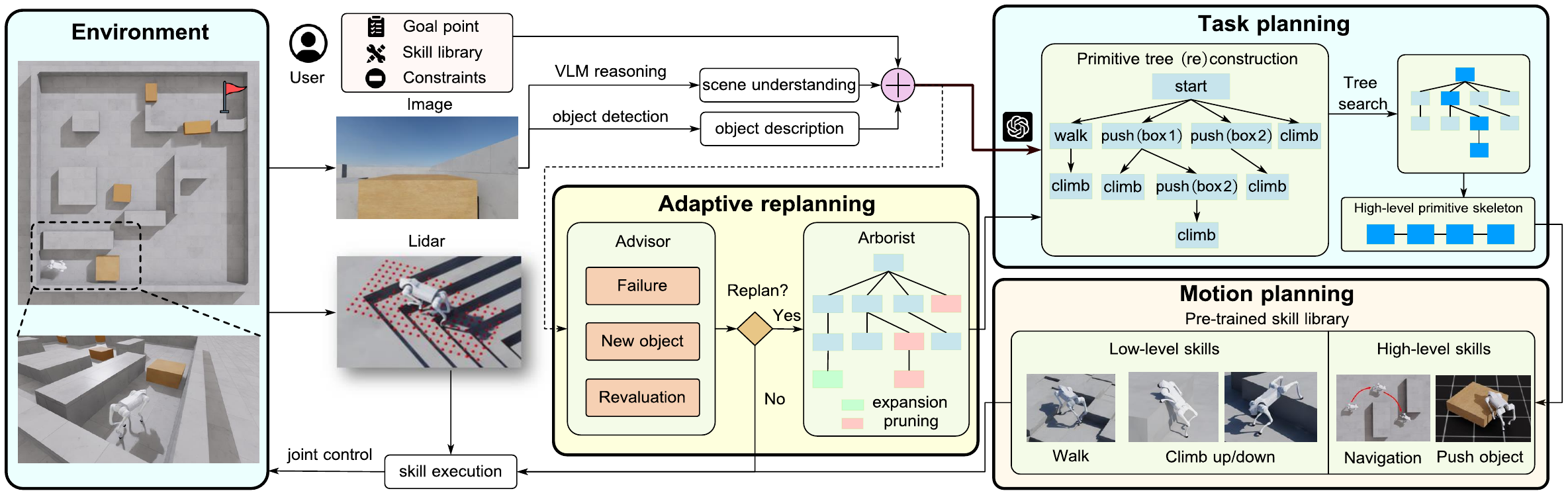} 
\caption{An overview of AINav approach, a hierarchical system for interactive navigation. 
The task planning module processes visual input to generate a subtask skeleton for execution by the motion planning module, integrated with an adaptive replanning mechanism that enables flexible replan triggering and rapid adjustments to the task plan.
}
\label{Fig: overview} 
\end{figure*}

\subsection{Closed-Loop Task Planning with LLMs}
Due to superior performance in language understanding and commonsense reasoning, LLMs are widely adopted to efficiently solve closed-loop task planning, which refers to generating a series of intermediate steps to achieve the specific goal while modifying the plan based on new observations.
Many works incorporate an iterative manner for closed-loop task planning, which replans based on the description or analysis of real-time observations~\cite{liu2023reflect,wang2024llm}.
However, these methods typically generate a single plan for execution, which often contains errors or infeasible decomposition due to model uncertainties. 
This increases the need for replanning and hinders computationally efficient planning.
Tree-based structure for LLM reasoning and planning enhances the solution quality by generating multiple plans and select the optimal tree path with state evaluation~\cite{yao2024tree, hu2023tree,zhao2024large}.
LLM-BT~\cite{zhou2024llm} combines LLM with behavior trees to automatically update the tree, addressing the reactive rigidity of traditional methods to adapt to environmental changes.
However, these methods either lack a replanning mechanism or trigger replanning solely upon encountering failures and concentrate primarily on error correction, overlooking a comprehensive understanding and effective utilization of interactive environments.

\begin{figure*}[!t] 
\centering
\includegraphics[width=0.92\linewidth]{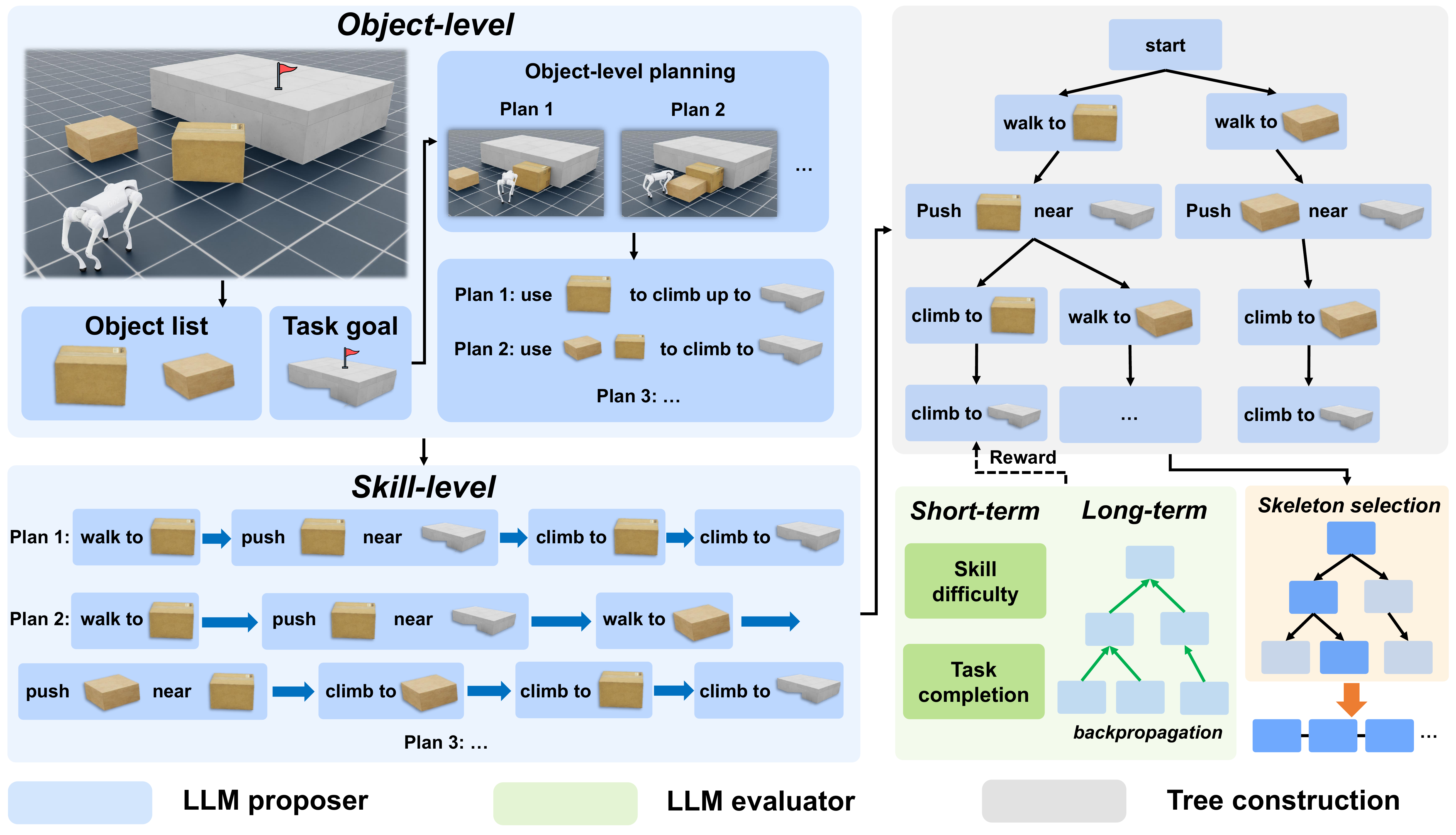} 
\caption{Task planning with the proposer and the evaluator.
}
\label{Fig: task_planning} 
\end{figure*}

\section{Problem Formulation}
We formalize interactive navigation as a 6-tuple TAMP problem $(\mathcal{X}_{init},\mathcal{X}_{goal},\mathcal{O}_s,\mathcal{O}_m,\mathcal{A}, \mathcal{C})$, with robot's initial position $\mathcal{X}_{init}$, navigation goal position $\mathcal{X}_{goal}$, fixed obstacles $\mathcal{O}_f$, movable obstacles $\mathcal{O}_m$, action space $\mathcal{A}$, and constraints $\mathcal{C}$.
The objective of interactive navigation is to execute a sequence of actions $\pi = [a_1, \ldots, a_n]$ that enables the robot to move from initial position $\mathcal{X}_{init}$ to goal position $\mathcal{X}_{goal}$, where each action $a_i =(p_i,q_i) \in \mathcal{A}$ ($i=1,\ldots,n$) consists of a primitive skill $p_i$ from the pre-defined skill library and continuous parameters $q_i$ for execution policy.
Finally, each action $a_i$ is executed by a pre-trained policy that generates the joint control commands for quadruped robot actuation.
We assume that the robot knows whether the objects within its field of view are movable, i.e., whether an object belongs to $\mathcal{O}_f$ or $\mathcal{O}_m$.
Unlike traditional navigation problems that assume a fixed free space and typically the existence of a feasible path to the goal, interactive navigation tasks consider scenarios where no feasible path exists in the original workspace, and the robot needs to proactively change the workspace through interactions to create feasible paths.
We propose a hierarchical planning framework, AINav, which decides the primitive skills to execute for task planning and specifies execution details for motion planning, jointly optimizing discrete skill selection and continuous execution parameters to tackle interactive navigation problems.

\section{Adaptive Interactive Navigation}

\subsection{Framework Overview}

AINav consists of three primary components: LLM-based task planning, motion planning with pre-trained skill library, and adaptive replanning, as shown in~\Cref{Fig: overview}.
The robot first captures the first-person perspective images, and extracts scene understanding (e.g., whether pathways are obstructed) using a VLM, and an object detector provides object descriptions including spatial positions and geometric properties.
By combining user instructions, including navigation goal, skill library, and constraints, with environmental information as inputs, the task planning module uses the LLM to construct a primitive skill tree for reasoning and task decomposition, enabling the robot to obtain the optimal high-level primitive skeleton.
Taking the height data from lidar as input, the motion planning module executes specific low-level actions following the skeleton by leveraging a pre-trained skill library consisting of robust locomotion and interaction skills. 
Finally, a novel adaptive replanning mechanism interprets new egocentric observations, not only determining whether replanning is necessary but also adjusting the current plan for swift adaptation to incremental environmental observations.
The detailed prompt of the proposed LLM-based proposer, evaluator, advisor, and arborist can be found in \Cref{appendix: prompts}.

\subsection{Task Planning with Primitive Skill Tree}

We propose the primitive skill tree for task planning shown in \Cref{Fig: task_planning}.
The primitive skill tree encompasses two LLM-based roles: a proposer for primitive tree construction denoted as $\text{LLM}_p$, and an evaluator for node evaluation and skeleton selection, denoted as $\text{LLM}_e$. 

\textbf{Proposer: object-level to skill-level.}
Given the user instructions $I$ and environmental information $O$, we first utilize $\text{LLM}_p$ to propose multiple potential object-level strategies $P_{obj}$, where each plan specifies which objects to interact with and how to leverage them to achieve the navigation goal,
\begin{equation}
\begin{split}
P_{obj} = \text{LLM}_p(I, O).
\end{split}
\label{eqn:object-level-plan}
\end{equation}
Building upon object-level plans, we subsequently generate standardized skill-level plans $P_{skill}$, where each step corresponds to a primitive skill from the robot's skill library $S$,
\begin{equation}
\begin{split}
P_{skill} = \text{LLM}_p(P_{obj}, S).
\end{split}
\label{eqn:skill-level-plan}
\end{equation}
Each primitive skill consists of the skill name and symbolic parameters, including walk (target position), climb (target position), navigate (target position), and push (object id, target position).
After skill-level plans are generated, we construct the primitive skill tree by merging steps of plans in $P_{skill}$ with identical historical traces.
Specifically, each plan path is iteratively added to the tree.
If a sequence of primitive skills and parameters already exists in the tree, new plans are merged with the existing structure at that common node. 
Otherwise, a new tree branch is created.
This approach offers two key advantages: First, unlike other approaches that use the LLM to directly generate a single task plan, multiple alternative plans proposed by the LLM are more robust to unexpected situations, such as misunderstandings by the LLM, limited scene observation, and environmental uncertainties.
Second, compared to directly using the LLM to generate detailed plans that may lead to illogical or inexecutable plans, transitioning from holistic object reasoning and utilization to specific skill planning tends to propose more reasonable and feasible interaction strategies.

\textbf{Evaluator: short-term to long-term.}
To evaluate the proposed plans, we utilize $\text{LLM}_e$ to assess the immediate reward $r(n)$ of each node $n$ in the primitive tree generated by proposer $\text{LLM}_p$, which considers the contribution to the specific task and the difficulty of pre-defined primitive skills.
For example, if the robot interacts with objects like pushing movable obstacles to create a free path for navigation, this skill is essential to completing the task, but the interaction is also difficult to complete.
If a skill is non-executable, the subsequent nodes including this node will be deleted to ensure the feasibility of the plan.
If a trajectory in the primitive tree reaches the navigation goal at the end, an additional bonus is given to the leaf node as the terminal reward.
Finally, we apply a backpropagation method to update the long-term cumulative reward $Q(n)$ of nodes as follows,
\begin{equation}
\begin{split}
Q(n) = r(n) + \gamma \cdot \frac{1}{|\text{children}(n)|} \sum_{n_c \in \text{children}(n)} Q(n_c),
\end{split}
\label{eqn:1}
\end{equation}
where $\gamma$ is the discount factor. 
By iteratively selecting the child nodes with maximal cumulative reward, we derive the optimal high-level primitive skeleton. 
The symbolic parameters are then converted into precise coordinates using the LLM for execution, which is detailed in \Cref{appendix: prompts}.
This tree-structured approach also allows the existence of alternative strategies, enabling rapid replanning upon receiving new environmental information, which will be described in \Cref{sec: replanning}.

\subsection{Skill Pre-training with Reinforcement Learning}

To robustly execute the decomposed tasks, we train a primitive skill library using RL for motion planning.
The skill library includes both low-level locomotion skills, such as walking and climbing, and high-level skills like navigation and object-pushing strategies.

In low-level locomotion skill training, the walking policy training is formulated as a velocity-tracking task, 
which encourages robots to follow the linear and angular velocity command while adapting to arbitrary terrain.
As for the climbing skill, to ensure the robot learns an effective policy, we refine the reward design to encourage the robot to reach the goal while aligning its movement direction toward the goal.
Specifically, we add a customized head collision penalty to prevent reliance on the head when climbing up high hurdles, encouraging the robot to use its legs as support instead of the head to learn a more natural and safe climbing action. 
Additionally, curriculum learning is employed to gradually increase terrain difficulty or hurdle height, enabling the development of robust locomotion policies.

Building upon low-level locomotion policies, we train high-level skills using the hierarchical RL method as a pose-tracking task, which generates velocity commands for the pre-trained locomotion policy to follow.
Specifically, the pushing skill aims to push boxes to the target pose, and the navigation skill navigates the robot to the target pose while avoiding obstacles.
We randomize the shape and size of the boxes for pushing skill learning and obstacle density for navigation skill acquisition, facilitating robust high-level skill learning.
All these skills are trained using PPO~\cite{schulman2017proximal} in the IsaacLab simulation environment.
The reward terms of all skills are detailed in~\Cref{appendix: reward-design}.
After training these primitive skills, 
we encapsulate them into APIs for task planning, enabling sequential skill execution to accomplish complex tasks.

\subsection{Adaptive Replanning with Observation Interpretation}
\label{sec: replanning}

Unlike previous LLM-based planning approaches that assume omniscient scene description, navigation tasks typically involve unknown environments, where the robot acquires new observations with an egocentric perspective.
For this reason, we develop the adaptive replanning mechanism to analyze the new observations and efficiently replan when needed. 
This mechanism consists of two LLM-based agents, an advisor and an arborist, designed to address two key questions: when to replan, and how to enhance replanning efficiency.

Taking new environmental observations and the current plan as inputs, the advisor conducts an observation interpretation and determines whether replanning is necessary.
We propose three aspects of observation interpretation, including failure, new objects, and revaluation.
Failure indicates the current plan is practically infeasible, new objects represent the perception of a new object that may be useful, and revaluation means reassessing the plan based on updated environmental information.
Unlike previous works that only replan when encountering a failure situation, the other two types of comprehension are also important to perform tasks with an incrementally updated understanding of the environment.
For example, upon detecting a new object, the robot needs to assess its geometric properties and determine whether utilizing it would enhance task performance.
As for revaluation, perceived object properties, such as whether movable, may change during the robot's movement, leading to updated affordances that may necessitate revaluating the interaction strategies in the current plan.
Triggered by one of these three interpretations, the advisor decides whether to replan by comparing potential improvements against the current plan.

If the advisor determines that replanning is necessary, it generates specific suggestions about adjustments to the current plan according to the interpretation of the new observations for the arborist, which subsequently modifies the primitive tree structure through node addition and pruning operations.
Node addition takes new objects into consideration in the primitive tree, while node pruning removes failed paths. 
This tree-based architecture enables computationally efficient replanning through its inherent structural flexibility, facilitating real-time task execution.
Following structural modifications, the system performs backpropagation operations within the updated tree and selects an optimal plan skeleton for execution using pre-trained skills.
Through the adaptive replanning approach, the robot can respond to new environmental information flexibly and adjust the planning strategies accordingly. 

\section{Simulation}
\label{headings}

\begin{figure}[!t] 
\centering
\includegraphics[width=0.9\linewidth]{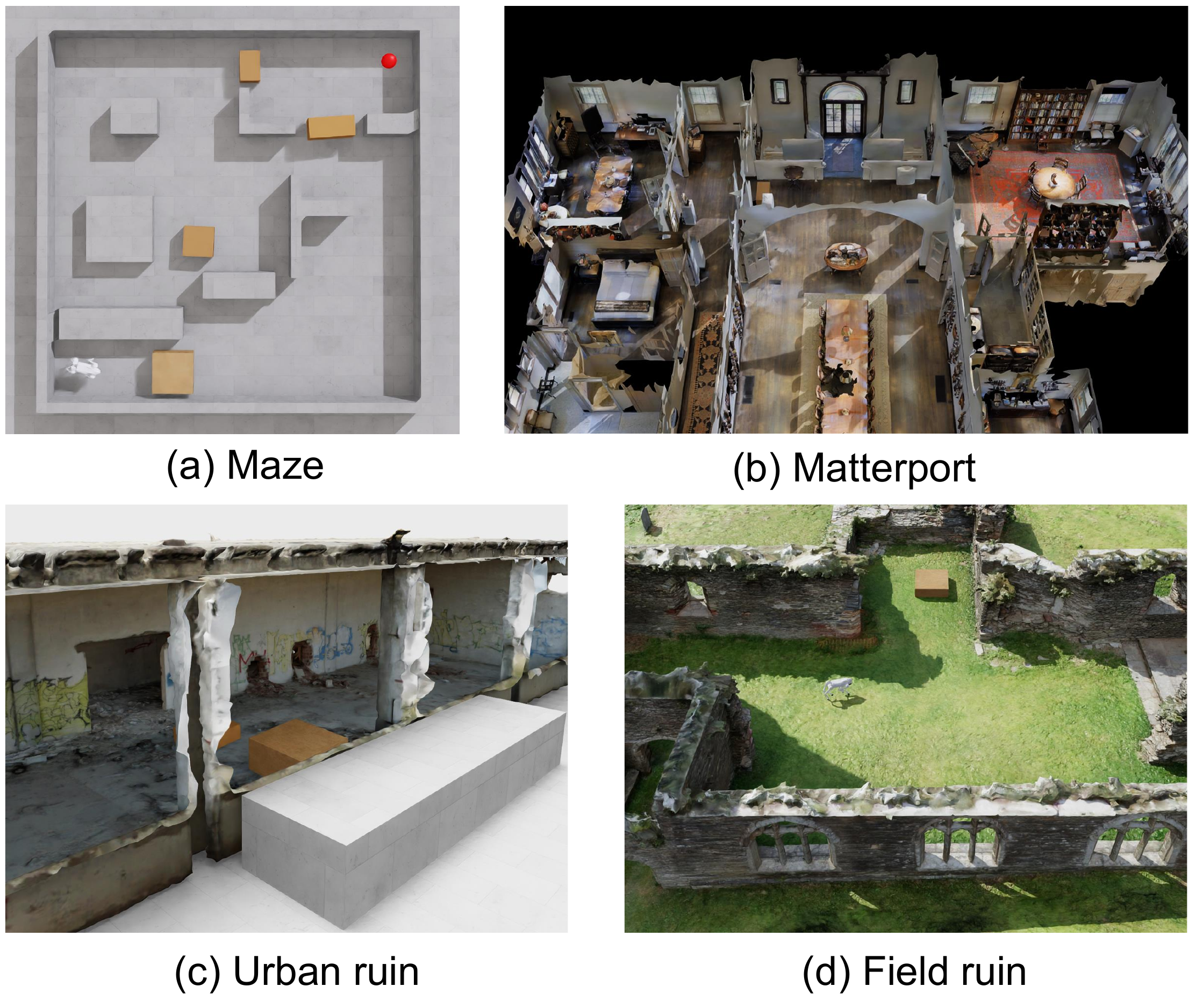} 
\caption{
Simulation environments for interactive navigation tasks. }
\label{Fig: simulation_environment} 
\end{figure}

\subsection{Experimental Setup}
\textbf{Simulation Settings.}
We conduct simulations in IsaacSim to validate the effectiveness of AINav. 
The simulation environment includes the maze, indoor Matterport3d dataset, urban ruin, and field ruin, as shown in \Cref{Fig: simulation_environment}. 
In the maze environment, we design diverse tasks and scenarios, which will be detailed in the following section.
For the other three environments, we test AINav in simulated realistic indoor and outdoor scenarios, including cluttered domestic environments and ruin scenes.
We use the GPT-4o model as the LLM backbone for all methods in our experiments, the CLIP model as VLM, the YOLOv8 model as the detection model, and the Unitree Go2 as the robotic platform to perform the simulations. 
To simplify visual processing, we assume that the robot can obtain noisy dimensions of objects within the robot's field of view, modeled as white noise with a standard deviation proportional to the distance between the robot and the objects.
All simulations are conducted on a desktop with an Intel Core i7 CPU @ 2.10 GHz and 16GB RAM.

\begin{table*}[t]
    \centering
    \caption{Performance metrics across different scenarios and methods}
    \footnotesize
    \begin{tabular}{cc ccc ccc ccc ccc}
        \toprule
        \multirow{2}{*}{Method} & \multirow{2}{*}{Metric} & \multicolumn{3}{c}{Box obstruction} & \multicolumn{3}{c}{Box usage} & \multicolumn{3}{c}{Stair building} & \multirow{2}{*}{Matterport} & \multirow{2}{*}{Urban ruin} & \multirow{2}{*}{Field ruin} \\
        \cmidrule(lr){3-5} \cmidrule(lr){6-8} \cmidrule(lr){9-11} 
        & & L & M & H & L & M & H & L & M & H &  &  &   \\
        \midrule
        \multirow{5}{*}{Art-planner} & SR $\uparrow$& \textbf{100\%} & 0\% & 0\% & 80\% & 0\% & 0\% & 90\% & 0\% & 0\% & 0\% & 0\% & 0\% \\
        & OT $\downarrow$& \textbf{33.3} & 120.0 & 120.0 & 50.2 & 120.0 & 120.0 & \textbf{44.0} & 120.0 & 120.0 & 120.0 & 120.0 & 120.0 \\
        & OTS $\downarrow$& \textbf{33.3} & - & - & \textbf{32.8} & - & - & \textbf{35.6} & - & - & - & - & - \\
        & PT $\downarrow$& 5.6 & - & - & \textbf{5.9} & - & - & \textbf{5.8} & - & - & - & - & - \\
        & ET $\downarrow$& \textbf{27.7} & - & - & 26.9 & - & - & \textbf{29.8} & - & - & - & - & - \\
        & TLS $\downarrow$& \textbf{14.8} & - & - & 17.4 & - & - & \textbf{16.8} & - & - & - & - & - \\
        \midrule
        \multirow{5}{*}{Hierarchical RL} & SR & 40\% & 0\% & 0\% & 0\% & 0\% & 0\% & 0\% & 0\% & 0\% & 0\% & 0\% & 0\% \\
        & OT & 87.4 & 120.0 & 120.0 & 120.0 & 120.0 & 120.0 & 120.0 & 120.0 & 120.0 & 120.0 & 120.0 & 120.0 \\
        & OTS & 38.5 & - & - & - & - & - & - & - & - & - & - & - \\
        & PT & \textbf{0.1} & - & - & - & - & - & - & - & - & - & - & - \\
        & ET & 38.4 & - & - & - & - & - & - & - & - & - & - & - \\
        & TLS & 24.1 & - & - & - & - & - & - & - & - & - & - & - \\
        \midrule
        \multirow{5}{*}{RoboTool} & SR & 90\% & 80\% & 30\% & \textbf{90\%} & \textbf{70\%} & 20\% & 80\% & 60\% & 0\% & 30\% & 0\% & 20\% \\
        & OT & 65.9 & 79.5 & 104.3 & 64.8 & 80.5 & 109.2 & 70.9 & 92.5 & 120.0 & 110.3 & 120.0 & 113.9 \\
        & OTS & 59.9 & 69.4 & \textbf{67.8} & 58.7 & 63.6 & \textbf{66.2} & 58.6 & 74.1 & - & 87.8 & - & 89.7 \\
        & PT & 25.6 & 27.2 & 24.9 & 28.3 & 26.4 & \textbf{28.8} & 26.1 & 29.3 & - & 23.5 & - & \textbf{28.4} \\
        & ET & 34.3 & 42.2 & 42.9 & 30.4 & 37.2 & \textbf{37.4} & 32.5 & 44.8 & - & 64.3 & - & 61.3 \\
        & TLS & 19.6 & 24.3 & 29.5 & 19.2 & 23.7 & 26.7 & 20.9 & 31.4 & - & 43.4 & - & 36.7 \\
        \midrule
        \multirow{5}{*}{Tree-planner} & SR & 90\% & 80\% & 60\% & \textbf{90\%} & 50\% & 40\% & 80\% & 40\% & 30\% & 50\% & 0\% & 0\% \\
        & OT & 49.7 & 65.6 & 89.5 & 53.4 & 92.9 & 104.9 & 72.9 & 100.6 & 115.1 & 103.6 & 120.0 & 120.0 \\
        & OTS & 41.9 & \textbf{52.0} & 69.1 & 46.0 & 65.8 & 82.2 & 61.1 & \textbf{71.5} & 103.6 & 87.2 & - & - \\
        & PT & 10.7 & \textbf{15.7} & \textbf{18.8} & 21.2 & \textbf{25.9} & 29.8 & 26.4 & \textbf{28.5} & \textbf{36.1} & 32.2 & - & - \\
        & ET & 31.2 & 36.3 & 50.3 & \textbf{24.8} & 39.9 & 52.4 & 34.7 & \textbf{43.0} & 67.5 & 55.0 & - & - \\
        & TLS & 19.1 & 22.8 & 34.6 & 16.7 & 28.5 & 34.0 & 22.3 & 24.9 & 46.1 & 38.4 & - & - \\
        \midrule
        \multirow{5}{*}{AINav (Ours)} & SR  & \textbf{100\%} & \textbf{100\%} & \textbf{90\%} & \textbf{90\%} & \textbf{70\%} & \textbf{60\%} & \textbf{90\%} & \textbf{80\%} & \textbf{50\%} & \textbf{90\%} & \textbf{30\%} & \textbf{50\%} \\
        & OT & 42.3 & \textbf{53.3} & \textbf{77.4} & \textbf{50.0} & \textbf{78.7} & \textbf{96.7} & 52.6 & \textbf{85.3} & \textbf{108.5} & \textbf{77.3} & \textbf{113.1} & \textbf{97.8} \\
        & OTS & 42.3 & 53.3 & 72.7 & 42.2 & \textbf{61.0} & 81.1 & 45.1 & 76.6 & \textbf{96.9} & \textbf{72.7} & \textbf{97.1} & \textbf{75.6} \\
        & PT & 14.1 & 19.4 & 30.1 & 15.7 & 26.4 & 36.2 & 14.8 & 29.7 & 40.2 & \textbf{19.5} & \textbf{28.2} & 30.8 \\
        & ET & 28.2 & \textbf{33.9} & \textbf{42.6} & 26.5 & \textbf{34.6} & 44.9 & 30.3 & 46.9 & \textbf{56.7} & \textbf{53.2} & \textbf{68.9} & \textbf{44.8} \\
        & TLS & 17.4 & \textbf{18.9} & \textbf{25.6} & \textbf{16.2} & \textbf{19.2} & \textbf{26.4} & 17.1 & \textbf{23.4} & \textbf{31.3} & \textbf{34.9} & \textbf{24.6} & \textbf{21.5} \\
        \bottomrule
    \end{tabular}
    \begin{tablenotes}
      \footnotesize
      \item L = Low difficulty, M = Medium difficulty, H = High difficulty, SR = Success rate, OT = Overall time, OTS = Overall time under successful trials, PT = Planning time, ET = Execution time, TLS = Trajectory length under successful trials
    \end{tablenotes}
    \label{tab:quantitative_result}
\end{table*}

\begin{figure*}[!t] 
\centering
\includegraphics[width=0.92\linewidth]{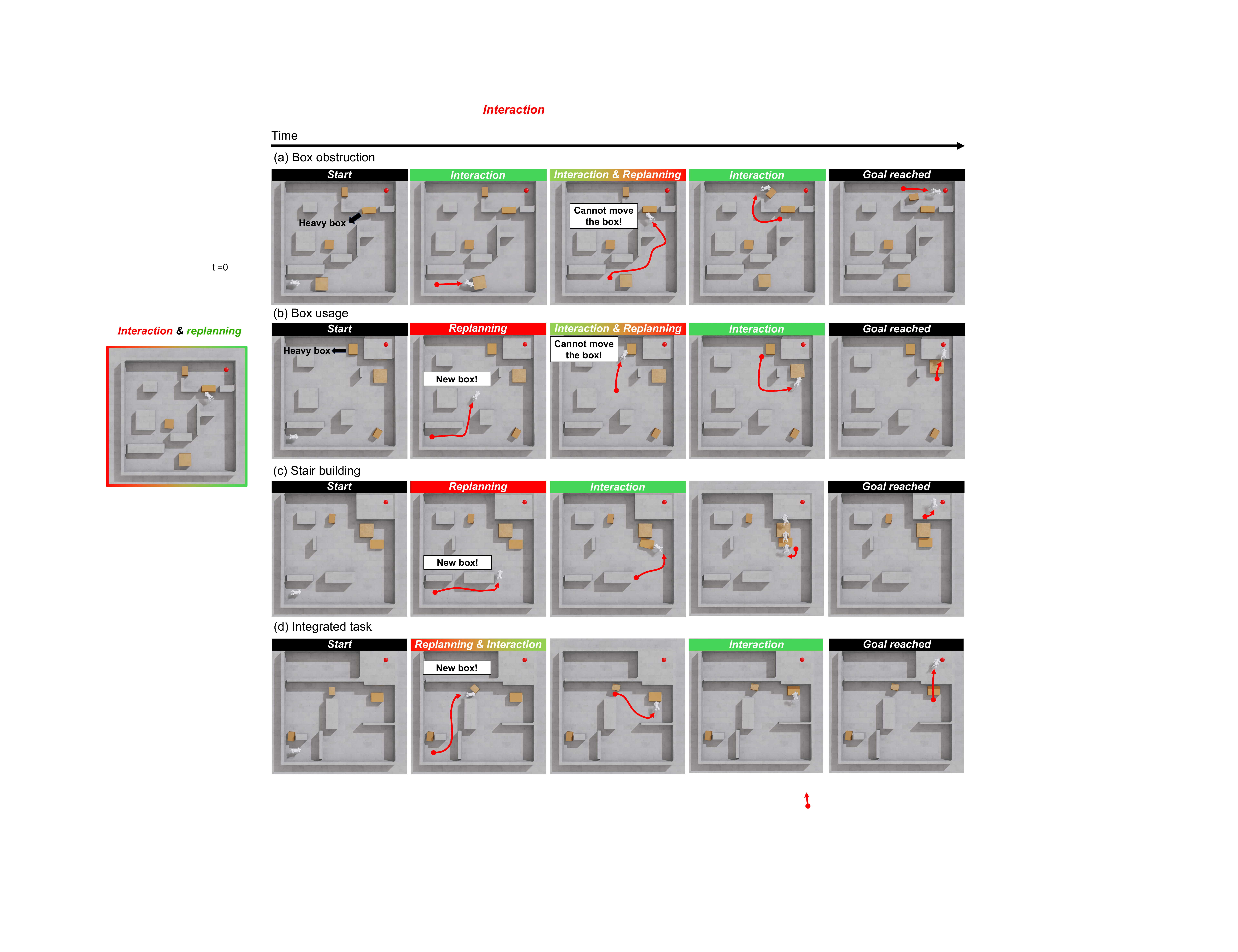} 
\caption{Success cases of AINav in maze environments. 
The green markers above the figures indicate the robot interacts with objects, while the red markers represent the robot performing replanning based on new environmental observations.
}
\label{Fig: qualitative_result} 
\end{figure*}

\begin{figure*}[!t] 
\centering
\includegraphics[width=0.95\linewidth]{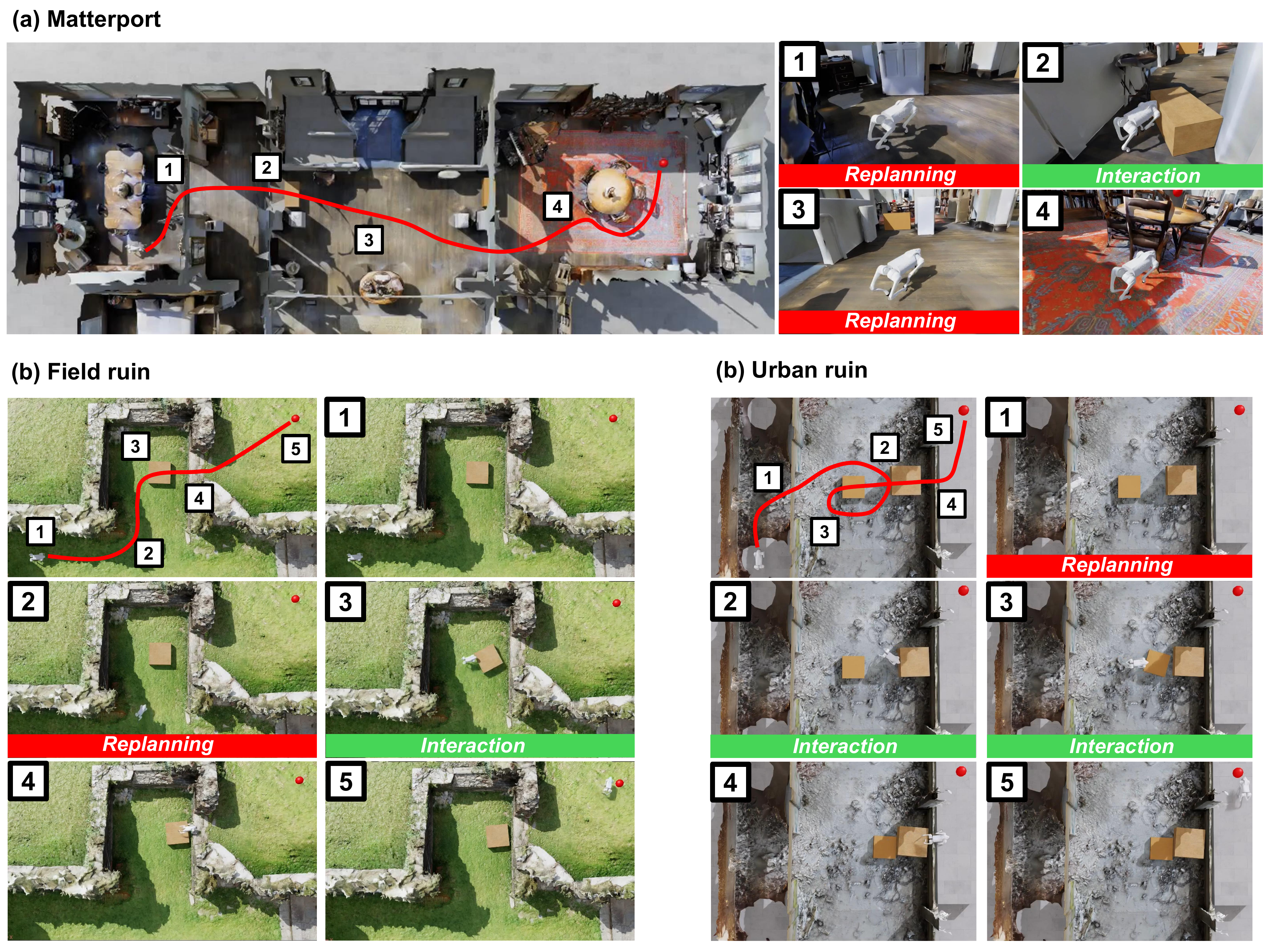} 
\caption{Success cases of AINav in Matterport3d, field ruin, and urban ruin. 
}
\label{Fig: qualitative_result2} 
\end{figure*}

\textbf{Task Description.}
In the maze environment, we design three interactive navigation tasks: box obstruction, box usage, and stair building, as illustrated in \Cref{Fig: qualitative_result}. 
The box obstruction task involves environments where boxes block the path.
Box usage and stair building require the robot to utilize boxes as tools to complete navigation tasks. 
Each scenario is divided into three levels of difficulty: low, medium, and high. 
The low-difficulty scenario removes the interaction tasks for simple navigation, which can be regarded as traditional navigation.
Medium difficulty requires an interaction, such as pushing movable obstacles or using a box as a tool, while high difficulty involves additional challenges with heavy boxes that are difficult to push, designed to evaluate the replanning performance and efficiency of various methods.
In the Matterport3d and ruin environments, we designate the start and navigation goal shown in \Cref{Fig: qualitative_result2} to test the effectiveness of AINav in diverse environments.
The robot needs to reach the designated goal, with a maximum climbing capability of $0.3$ meters as a constraint.
We set the maximum simulation time to $120$s and conducted $10$ repeated trials for each scenario.
A trial is considered successful if the robot achieves the goal within $120$s. 
Otherwise, the trial is considered a failure.

\textbf{Evaluation Metrics.}
The evaluation metrics include success rate (SR), overall time (OT), overall time under successful trials (OTS), planning time (PT), execution time (ET), 
and trajectory length under successful trials (TLS) to comprehensively assess the performance of different methods.
SR refers to the portion of successful trials across 10 trials. 
OT denotes the average simulation time across all trials. 
If a trial fails, OT is recorded as the maximum simulation time, which is $120$s.
OTS represents the average task completion time for successful trials. 
PT and ET indicate the average time spent on task planning and skill execution under successful completion, respectively.
TLS refers to the average trajectory length of successful trials.
SR, OT, OTS, and TLS indicate the overall effectiveness and efficiency of methods, while PT and ET reflect the computational efficiency and quality of task planning.

\begin{figure}[!t] 
\centering
\includegraphics[width=\linewidth]{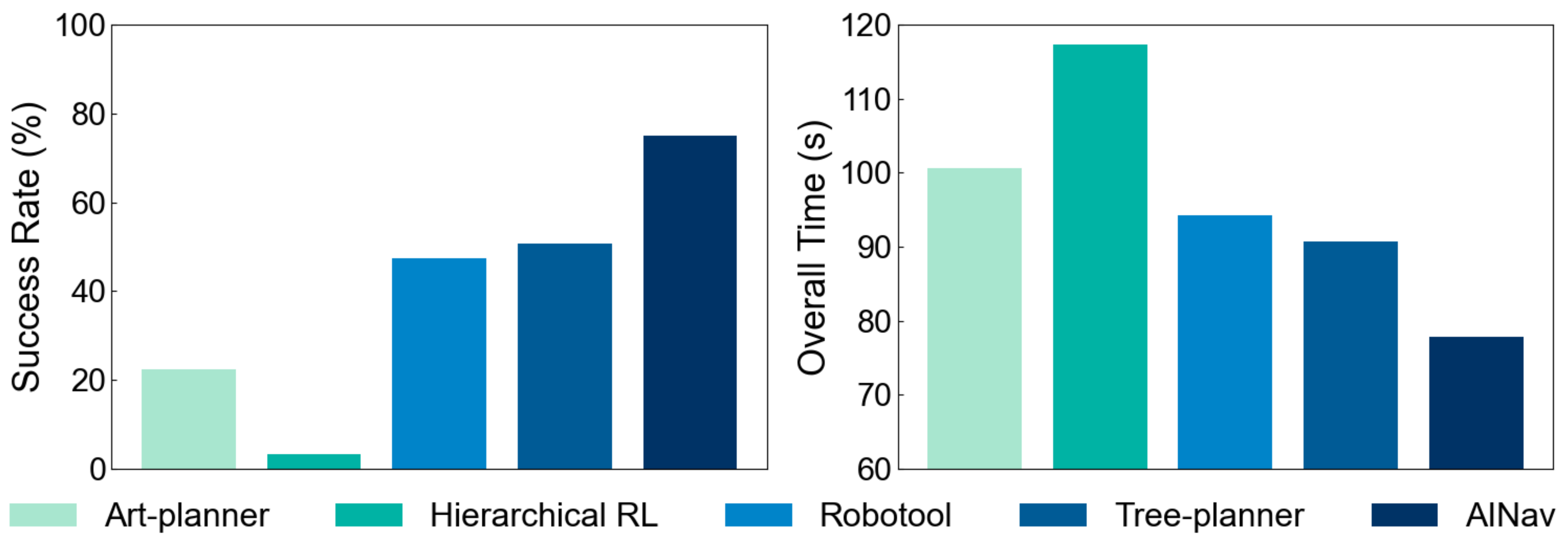} 
\caption{Average success rate and overall time across all scenarios of different methods. 
}
\label{Fig: quantitative_result} 
\end{figure}

\subsection{Baseline Methods}
We compare the performance of AINav to the following baselines.
\textbf{Art-planner}~\cite{wellhausen2023artplanner} is a sampling-based approach with learned motion cost for quadruped robots, chosen as a representative approach to solve traditional navigation tasks.
\textbf{Hierarchical RL} trains a high-level policy to function as the task planning module, which decides the low-level skill to use and the corresponding parameters to reach the goal.
We also select two LLM-based method for comparison.
The first is \textbf{RoboTool}~\cite{xu2023creative}, which employs multiple LLMs to perform different roles and invoke low-level skills to utilize tools, thereby solving complex planning and reasoning challenges. 
Since this method uses an open-loop formulation without reaction to new objects, we provide the complete environment description as its input and execute the skills suggested by its output.
The second method is \textbf{Tree-planner}~\cite{hu2023tree}, which employs a tree-structured reasoning approach to solve closed-loop TAMP problems. 
Because its plan sampling module requires global information, we include the initial object information as input.
We adjust the prompts used in these two LLM-based methods for better adaptation to the interactive navigation tasks.
For a fair comparison, all methods use the same set of low-level skills as AINav.

\subsection{Results and Analyses}

Simulation results are displayed in \Cref{tab:quantitative_result}.
We can observe that AINav exhibits higher success rates and less overall time compared to other baselines in diverse scenarios. 
The Art-planner achieves high task success rates in low-difficulty scenarios of the maze environment with low execution time, showcasing its effective planning capabilities in traditional navigation tasks. 
However, in other scenarios that require interaction, Art-planner is limited to passively adapting to the current environment without the capability to actively alter the workspace, thereby failing to accomplish more complex interactive navigation tasks.
Hierarchical RL can achieve limited success in low-difficulty scenarios, while in more intricate tasks that require interaction to solve, hierarchical RL struggles to identify the precise sequence of skills necessary for task completion in the expansive search space.

Leveraging the scene understanding and reasoning capabilities of pre-trained LLM, RoboTool and tree-planner achieve a higher success rate in interactive navigation tasks compared to Art-planner.
However, RoboTool is constrained by its open-loop planning scheme, which prevents it from adapting to new environmental information, resulting in lower success rates in high-difficulty tasks and disaster scenes.
While tree-planner leverages a closed-loop design to adjust its decisions based on new observations, its performance is highly dependent on the quality of the sampled plans, resulting in instability in planning results. 
Specifically, once the tree structure is generated, its inflexibility can hinder the planner from obtaining the optimal solution that does not exist in the original tree or changing to an updated better strategy in response to new observations.
Additionally, both methods require omniscient environmental information to generate task decomposition plans, rendering them ineffective in unknown environments.
In contrast, AINav not only enhances the quality of task planning outcomes through the design of a proposer and evaluator but also leverages adaptive replanning to flexibly adapt to new environmental information.
This ensures efficient planning and real-time responsiveness in unfamiliar environments without global environmental information as prior, resulting in high success rates in various scenarios. 
Although node evaluation slightly increases planning time, the improved task planning quality and flexible replanning mechanism enable AINav to achieve higher success rates, shorter overall completion time, and trajectory length compared to all baseline approaches.
As shown in \Cref{Fig: quantitative_result}, AINav demonstrates the highest success rate and the lowest overall completion time among all approaches and across various scenarios, showcasing its superior effectiveness in complex navigation tasks.

We visualize the performance of AINav across different tasks and scenarios in \Cref{Fig: qualitative_result} and \Cref{Fig: qualitative_result2}. 
In the box obstruction and Matterport scenario, the robot demonstrates the ability to comprehend the blocked environment, move boxes to create free space to reach the target position. 
In the box usage, stair building, and ruin scenarios, the robot actively utilizes movable obstacles to construct stairs, enabling access to higher areas, which are unreachable for a quadruped robot under normal circumstances.
Notably, AINav's advanced reasoning is highlighted in the stair-building and urban ruin tasks, where the robot must sequentially arrange boxes of varying heights in the correct order to climb to the prohibitively high platform. 
The successful completion of these tasks demonstrates the capacity of AINav for complex tool-use reasoning and execution.
Furthermore, we can observe that the robot can respond to updated information, such as when the box fails to move or the robot identifies new objects and tasks.
By leveraging an efficient closed-loop system, AINav enhances the robot's capability to adapt to incremental environmental comprehension, thereby facilitating effective task completion.

\begin{table}[t]
    \centering
    \caption{{Failure study of AINav in simulations}}
    \small
    \begin{tabular}{@{}ccc@{}}
        \toprule
        Primary Category & Sub-Category & Rate \\ \midrule
        \multirow{3}{*}{Task Timeouts} & Scene understanding error & $17.2\%$  \\
        & Incorrect task decomposition & $34.5\%$ \\
        & Skill execution unfinished & $24.1\%$  \\ 
        \midrule
        \multirow{2}{*}{Irrecoverable execution} & Intra-skill failure  &  $10.3\%$ \\ 
        & Skill transition failure &  $13.8\%$  \\
        \bottomrule
    \end{tabular}
    \label{tab:failure_study}
\end{table}

We further conduct a failure study to analyze the current limitations of AINav.
We categorize all failures observed during simulations into two primary types: task timeouts and irrecoverable execution failures, where task timeouts occur when the robot does not complete the task within the allocated time limit, and irrecoverable execution failures involve a critical error during primitive skill execution, such as the robot falling over.
\Cref{tab:failure_study} summarizes the distribution of these failure modes across simulations.
In summary, this failure study reveals that the primary bottlenecks for AINav lie in the efficiency of task planning and the seamless integration of low-level skills, highlighting clear directions for future improvement.
Furthermore, the assumption of simplified visual processing and reliance on predefined constraints to determine skill executability limit AINav's generalization in realistic scenarios, which will be the focus for future work to enhance robustness and adaptability.

\subsection{Ablation Studies}

\begin{figure}[!t] 
\centering
\includegraphics[width=0.9\linewidth]{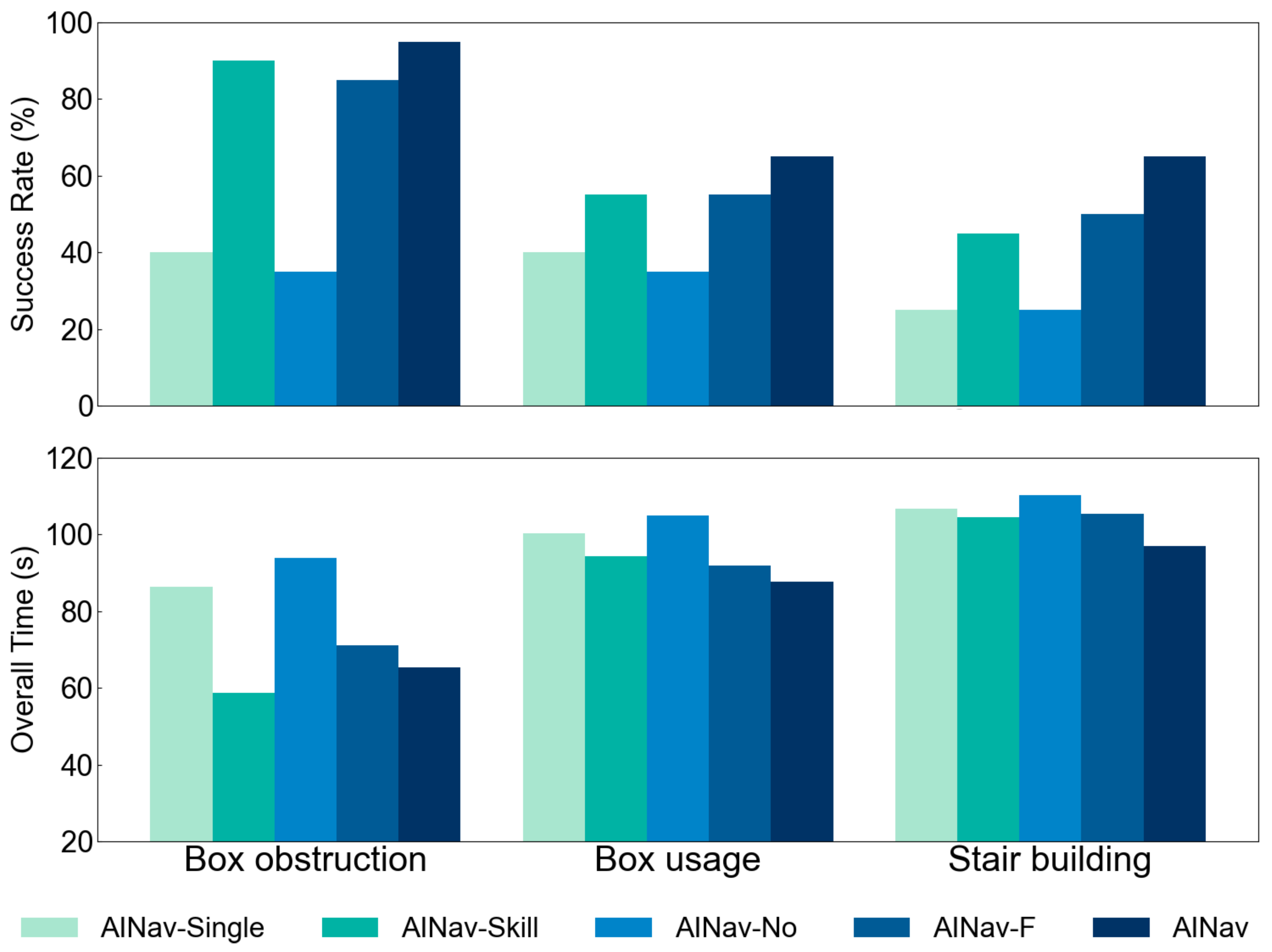} 
\caption{Comparison results in ablation study. 
}
\label{Fig: ablation_study} 
\end{figure}

We conduct ablation studies in maze environments to compare AINav with the following four variants:
1) \textbf{AINav-Single} that only generates a single plan with LLM in the task planning module;
2) \textbf{AINav-Skill} that directly generates skills without incorporating object-centric reasoning;
3) \textbf{AINav-No} that removes the replanning module;
4) \textbf{AINav-F} that only replans when a failure is encountered.
We evaluate these methods in the middle-difficulty and high-difficulty scenarios of box obstruction, box usage, and stair building in the maze environment, which need interaction for task completion, to compare the SR and OT.
The results of ablation studies are presented in \Cref{Fig: ablation_study}.
AINav-Single exhibits low success rates, as the single plan generated by the LLM is often flawed or inefficient due to inherent uncertainties of LLM, leading to task failures.
Benefiting from the tree structure that enhances the robustness and planning quality, AINav-Skill achieves a reasonable success rate. 
However, it lacks a holistic understanding of object utilization in complex scenarios, leading to inexecutable task decomposition plans. 
AINav-No performs poorly because it lacks a replanning phase, making it unable to adapt to new observations. 
Conversely, AINav-F improves task success rates through failure recovery but tends to miss new objects that could offer more optimal solutions, thus often resulting in excessive time spent and task timeouts. 
By integrating tree-structured reasoning with the adaptive replanning module, AINav enhances its task-planning capabilities and adaptability to incremental environmental information, achieving the highest success rates and improved time efficiency.

\section{Real-World Experiment}

\begin{figure*}[!t] 
\centering
\includegraphics[width=\linewidth]{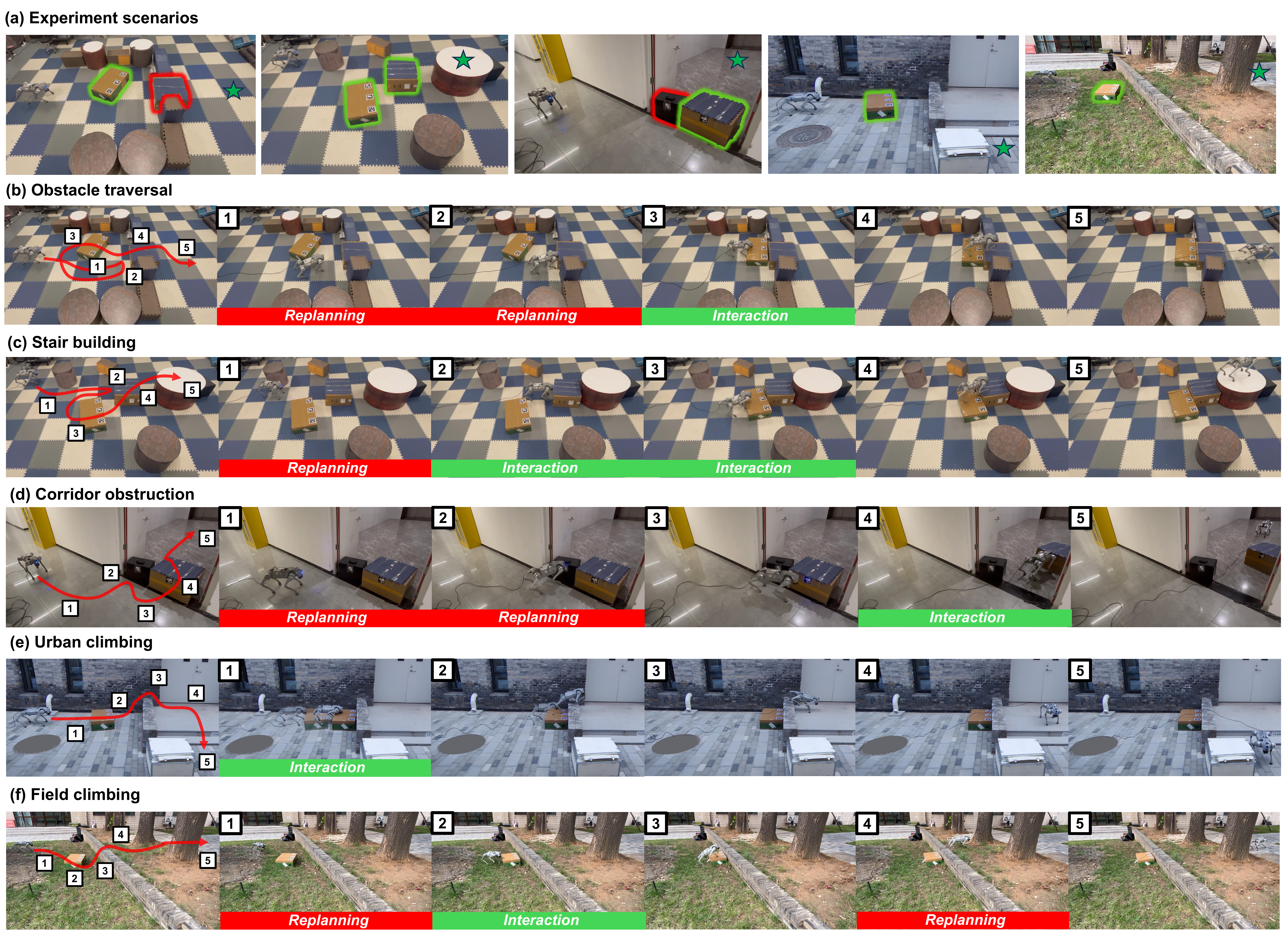} 
\caption{Visualization of real-world experiments.
(a) shows various scenarios in real-world experiments, where green stars indicate the navigation goal.
Highlighted boxes indicate movable objects, with green markers denoting light movable boxes and red markers denoting heavy boxes.
(b)-(f) illustrate case studies for each scenario, in which the leftmost subfigure shows the overall robot trajectory, while the subsequent key frames detail the specific process of reaching the navigation goals.
}
\label{Fig: real_exp}
\end{figure*}

We conduct real-world experiments to validate the performance of AINav in practical situations.
We employ the Unitree Go2 as the robotic platform, which is equipped with an L1 Lidar for terrain information acquisition and an onboard RGB camera for visual detection.
We evaluate the effectiveness of AINav across five interactive navigation scenarios, as illustrated in \Cref{Fig: real_exp} (a).
The first three scenarios are conducted in indoor environments: obstacle traversal, stair building, and corridor obstruction. 
In obstacle traversal, the robot is required to utilize a box as a stepping stone to surmount a $0.35$m obstacle. 
The stair-building task requires the robot to arrange boxes of two different heights to construct a temporary staircase, enabling it to ascend a $0.50$m platform. 
In corridor obstruction, two boxes obstruct a narrow passageway, one of which is heavy to push. 
The robot must proactively interact with these obstacles to create a navigable path and replan its strategy upon encountering the fixed box. 
The final two scenarios are set in outdoor environments, where the robot needs to utilize a box to traverse a $0.45$m obstacle on the concrete floor or grass.
We assume the quadruped’s maximum climbing height is $0.30$m.
We benchmark our method against the RoboTool baseline, which performs reasonably well in our simulations and has been validated through real-world experiments~\cite{xu2023creative}.
We adopt the SLAM Toolbox package to obtain the robot's real-time pose.
Focusing primarily on the reasoning and planning capabilities of AINav, we utilize Apriltag detection to obtain object pose and dimension, thereby eliminating the risk of confounding factor detection errors.

\begin{figure}[!t] 
\centering
\includegraphics[width=\linewidth]{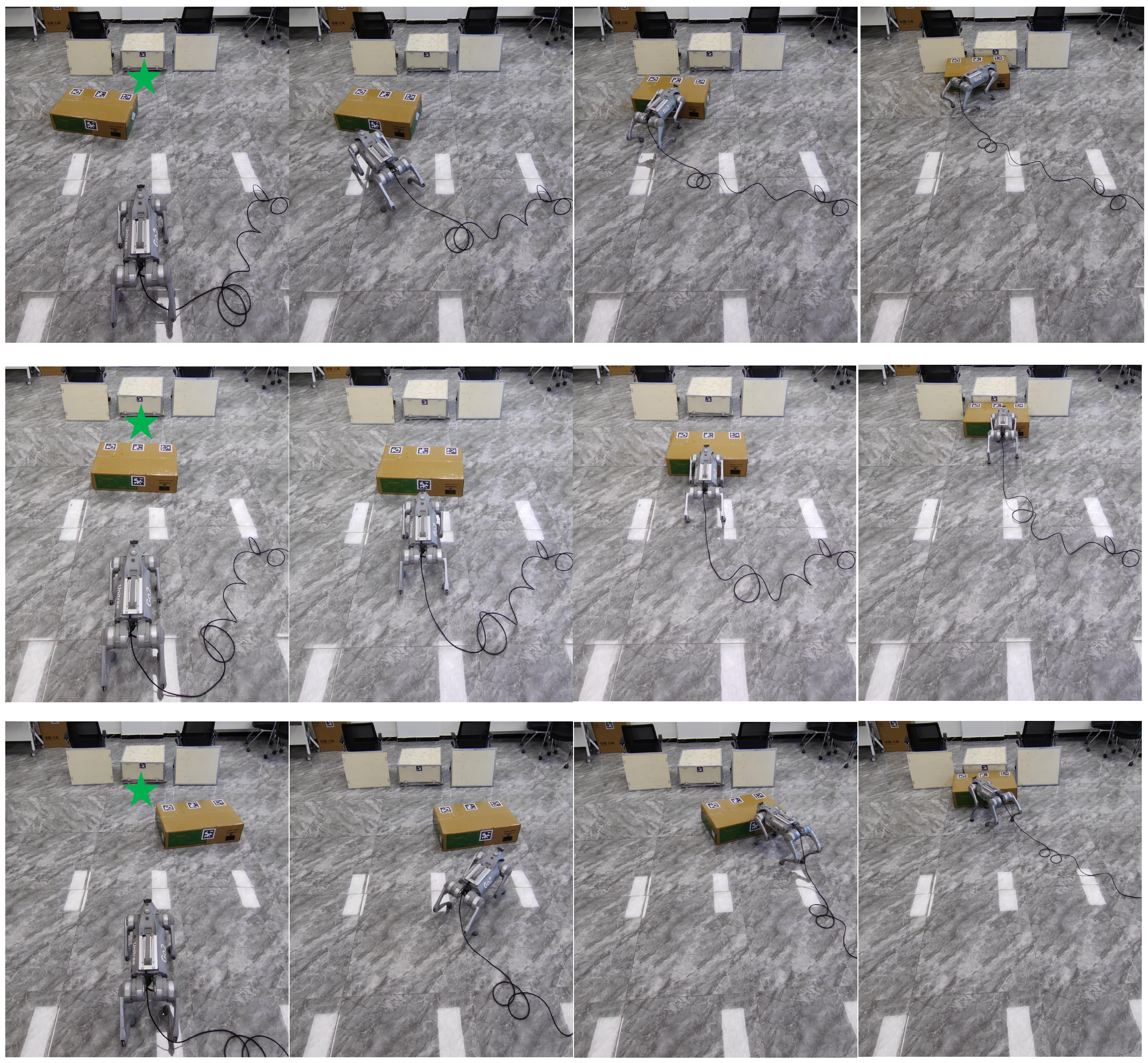} 
\caption{Box-pushing policy in different configurations. 
Each row demonstrates the pushing strategy under different initial box positions: left-shifted (top), centered (middle), and right-shifted (bottom).
}
\label{Fig: box_pushing} 
\end{figure}

\begin{table*}[t]
    \centering
    \caption{Metrics comparison in real-world experiments}
    \small
    \begin{tabular}{@{}cccccccccc@{}}
        \toprule
        \multirow{2}{*}{Scenario} & \multicolumn{4}{c}{RoboTool} & \multicolumn{4}{c}{AINav} \\
        \cmidrule(lr){2-5} \cmidrule(lr){6-9}
         & SR & PT (s) & ET (s) & TLS (m) & SR & PT (s) & ET (s) & TLS (m) \\
        \midrule
        Obstacle traversal  & $20\%$ & 16.4 & 40.2 & \textbf{6.8} & $\textbf{80\%}$ & \textbf{11.8} & \textbf{38.4} & 8.1 \\
        Stair building      & $0\%$ & - & - & - & $\textbf{20\%}$ & \textbf{16.5} & \textbf{71.6} & \textbf{9.3} \\
        Corridor obstruction& $40\%$ & \textbf{15.9} & 58.4 & 6.9 & $\textbf{80\%}$ & 18.1 & \textbf{53.7} & \textbf{6.4} \\
        Urban climbing      & $60\%$ & 14.3 & 45.2 & 7.0 & $\textbf{80\%}$ & \textbf{8.4} & \textbf{37.6} & \textbf{6.6} \\
        Field climbing      & $\textbf{60\%}$ & 19.2 & 62.4 & 12.3 & $\textbf{60\%}$ & \textbf{11.3} & \textbf{60.4} & \textbf{9.4} \\
        \midrule
        Average success rate            & \multicolumn{4}{c}{$36\%$} & \multicolumn{4}{c}{$\textbf{64\%}$} \\
        \bottomrule
    \end{tabular}
    \label{tab:exp_sr}
\end{table*}

\Cref{Fig: real_exp} (b) showcases AINav's capacity for dynamic replanning and creative tool utilization. Initially, the robot attempts a direct pushing strategy to withdraw a box blocking its path. 
Upon discovering that the box (red highlighted in \Cref{Fig: real_exp} (a)) is difficult to push, AINav replans and pushes a nearby smaller box as a support structure to scale the obstacle. 
This adaptation from unpredictable environmental information underscores AINav's ability to traverse impassable barriers through tool use and robust strategic replanning.
\Cref{Fig: real_exp} (c) showcases AINav's advanced tool utilization capability. 
The robot constructs a makeshift staircase from boxes of varying heights to ascend a 0.5m platform, which is twice the height of the robot itself, showing AINav's long-horizon reasoning and planning abilities.
Although the task planner consistently generates viable solutions, the relatively low success rate is primarily attributed to execution failures, particularly during the challenging task of sequential climbing on non-fixed boxes. 
Addressing this execution limitation will be a key focus of our future work.
\Cref{Fig: real_exp} (d) demonstrates the experiment on corridor obstruction. 
After an initial attempt to push one of two obstructive boxes fails, the robot strategically interacts with the second box, clearing a navigable path to its destination.
\Cref{Fig: real_exp} (e) and (f) show the robot completing the interactive navigation task in the outdoor scenarios, employing task decomposition and skill execution to achieve the navigation goal, overcoming challenges that traditional navigation methods could not address. 
Note that although the friction between indoor and outdoor scenarios is different, the robot can still push the box to the destination, demonstrating the robustness of the pre-trained pushing skill for interaction.
These real-world experiments demonstrate AINav's capability to perform interactive navigation tasks across diverse scenarios, highlighting its proficiency in both complex tool utilization and dynamic replanning in response to new environmental information. 
Further details are available in our supplementary video.
\Cref{tab:exp_sr} presents the metric comparison across different scenarios in $5$ repeated experiments, where AINav achieves a higher success rate, less planning time, and execution time, highlighting the effectiveness and adaptivity of AINav in various tasks and both indoor and outdoor settings.

To evaluate the robustness of the pushing skill, we also carry out box-pushing tests with varying initial box positions in real-world scenarios, as shown in \Cref{Fig: box_pushing}. 
We observe that the robot consistently pushes the box to the target position under different conditions, regardless of changes in the initial box placement.
This stable performance across diverse configurations demonstrates the robustness of the pushing skills, facilitating effective interaction for navigation task completion.

\section{Conclusion}

We introduce an LLM-based adaptive interactive navigation approach, AINav, which proactively utilizes interactive objects to create feasible paths to reach originally unreachable goals. 
Our hierarchical framework leverages the LLM for task planning and employs pre-trained skills with RL for motion planning.
The tree-structured task decomposition not only enhances the planning quality for long-horizon complex tasks but also allows for autonomous modifications in specific tree parts when new observations are obtained.
Leveraging RL-driven pre-trained robust locomotion and interaction skills, the quadruped robot can manipulate the environment and create feasible paths to reach the navigation goal.
Moreover, an adaptive replanning method is presented for intelligent understanding of incremental information, enabling robots to quickly adapt under partially observable situations.
AINav has the potential to enhance robotic navigation capabilities, offering an effective and efficient solution for interactive navigation in challenging environments.

Future work will work on tighter integration between foundation models and versatile motion capabilities for real-time embodied planning and control. 
We plan to leverage visual foundation models to identify task-related objects and extract corresponding properties, broadening the practical applicability of AINav to real-world scenarios.
Moreover, we will expand the variety of primitive skills and enhance their robustness and interior seamless transition, ultimately aiming to extend our framework to a diverse range of robotic platforms and complex, long-horizon tasks.


{
\bibliographystyle{ieeetr}
\bibliography{ref}

@string {IROS = "IROS"}

@string {ICRA = "ICRA"}

@string {IROS = "IEEE/RSJ International Conference on Intelligent Robots and Systems"}

@string {ICRA = "IEEE International Conference on Robotics and Automation"}

@article{hoeller2024anymal,
  title={Anymal parkour: Learning agile navigation for quadrupedal robots},
  author={Hoeller, David and Rudin, Nikita and Sako, Dhionis and Hutter, Marco},
  journal={Science Robotics},
  volume={9},
  number={88},
  pages={eadi7566},
  year={2024},
  publisher={American Association for the Advancement of Science}
}

@article{yao2024tree,
  title={Tree of thoughts: Deliberate problem solving with large language models},
  author={Yao, Shunyu and Yu, Dian and Zhao, Jeffrey and Shafran, Izhak and Griffiths, Tom and Cao, Yuan and Narasimhan, Karthik},
  journal={Advances in Neural Information Processing Systems},
  volume={36},
  year={2024}
}

@article{hu2023tree,
  title={Tree-planner: Efficient close-loop task planning with large language models},
  author={Hu, Mengkang and Mu, Yao and Yu, Xinmiao and Ding, Mingyu and Wu, Shiguang and Shao, Wenqi and Chen, Qiguang and Wang, Bin and Qiao, Yu and Luo, Ping},
  journal={arXiv preprint arXiv:2310.08582},
  year={2023}
}

@article{xu2023creative,
  title={Creative robot tool use with large language models},
  author={Xu, Mengdi and Huang, Peide and Yu, Wenhao and Liu, Shiqi and Zhang, Xilun and Niu, Yaru and Zhang, Tingnan and Xia, Fei and Tan, Jie and Zhao, Ding},
  journal={arXiv preprint arXiv:2310.13065},
  year={2023}
}

@article{ouyang2024long,
  title={Long-horizon Locomotion and Manipulation on a Quadrupedal Robot with Large Language Models},
  author={Ouyang, Yutao and Li, Jinhan and Li, Yunfei and Li, Zhongyu and Yu, Chao and Sreenath, Koushil and Wu, Yi},
  journal={arXiv preprint arXiv:2404.05291},
  year={2024}
}

@article{schulman2017proximal,
  title={Proximal policy optimization algorithms},
  author={Schulman, John and Wolski, Filip and Dhariwal, Prafulla and Radford, Alec and Klimov, Oleg},
  journal={arXiv preprint arXiv:1707.06347},
  year={2017}
}

@article{wellhausen2023artplanner,
  title={Artplanner: Robust legged robot navigation in the field},
  author={Wellhausen, Lorenz and Hutter, Marco},
  journal={arXiv preprint arXiv:2303.01420},
  year={2023}
}

@article{zhu2024cross,
  title={Cross Anything: General Quadruped Robot Navigation through Complex Terrains},
  author={Zhu, Shaoting and Li, Derun and Liu, Yong and Xu, Ningyi and Zhao, Hang},
  journal={arXiv preprint arXiv:2407.16412},
  year={2024}
}

@article{shah2023vint,
  title={ViNT: A foundation model for visual navigation},
  author={Shah, Dhruv and Sridhar, Ajay and Dashora, Nitish and Stachowicz, Kyle and Black, Kevin and Hirose, Noriaki and Levine, Sergey},
  journal={arXiv preprint arXiv:2306.14846},
  year={2023}
}

@inproceedings{zhang2024interactive,
  title={Interactive navigation in environments with traversable obstacles using large language and vision-language models},
  author={Zhang, Zhen and Lin, Anran and Wong, Chun Wai and Chu, Xiangyu and Dou, Qi and Au, KW Samuel},
  booktitle={2024 IEEE International Conference on Robotics and Automation (ICRA)},
  pages={7867--7873},
  year={2024},
  organization={IEEE}
}

@inproceedings{rudin2022advanced,
  title={Advanced skills by learning locomotion and local navigation end-to-end},
  author={Rudin, Nikita and Hoeller, David and Bjelonic, Marko and Hutter, Marco},
  booktitle={2022 IEEE/RSJ International Conference on Intelligent Robots and Systems (IROS)},
  pages={2497--2503},
  year={2022},
  organization={IEEE}
}

@article{wang2023describe,
  title={Describe, explain, plan and select: Interactive planning with large language models enables open-world multi-task agents},
  author={Wang, Zihao and Cai, Shaofei and Chen, Guanzhou and Liu, Anji and Ma, Xiaojian and Liang, Yitao},
  journal={arXiv preprint arXiv:2302.01560},
  year={2023}
}

@article{liu2023reflect,
  title={Reflect: Summarizing robot experiences for failure explanation and correction},
  author={Liu, Zeyi and Bahety, Arpit and Song, Shuran},
  journal={arXiv preprint arXiv:2306.15724},
  year={2023}
}

@article{zhao2024large,
  title={Large language models as commonsense knowledge for large-scale task planning},
  author={Zhao, Zirui and Lee, Wee Sun and Hsu, David},
  journal={Advances in Neural Information Processing Systems},
  volume={36},
  year={2024}
}

@inproceedings{cheng2024extreme,
  title={Extreme parkour with legged robots},
  author={Cheng, Xuxin and Shi, Kexin and Agarwal, Ananye and Pathak, Deepak},
  booktitle={2024 IEEE International Conference on Robotics and Automation (ICRA)},
  pages={11443--11450},
  year={2024},
  organization={IEEE}
}

@article{han2024lifelike,
  title={Lifelike agility and play in quadrupedal robots using reinforcement learning and generative pre-trained models},
  author={Han, Lei and Zhu, Qingxu and Sheng, Jiapeng and Zhang, Chong and Li, Tingguang and Zhang, Yizheng and Zhang, He and Liu, Yuzhen and Zhou, Cheng and Zhao, Rui and others},
  journal={Nature Machine Intelligence},
  pages={1--12},
  year={2024},
  publisher={Nature Publishing Group UK London}
}

@article{wang2024llm,
  title={LLM\^{} 3: Large Language Model-based Task and Motion Planning with Motion Failure Reasoning},
  author={Wang, Shu and Han, Muzhi and Jiao, Ziyuan and Zhang, Zeyu and Wu, Ying Nian and Zhu, Song-Chun and Liu, Hangxin},
  journal={arXiv preprint arXiv:2403.11552},
  year={2024}
}

@article{armleder2024tactile,
  title={Tactile-Based Negotiation of Unknown Objects during Navigation in Unstructured Environments with Movable Obstacles},
  author={Armleder, Simon and Dean-Leon, Emmanuel and Bergner, Florian and Guadarrama Olvera, Julio Rogelio and Cheng, Gordon},
  journal={Advanced Intelligent Systems},
  volume={6},
  number={3},
  pages={2300621},
  year={2024},
  publisher={Wiley Online Library}
}

@inproceedings{muguira2023visibility,
  title={Visibility-aware navigation among movable obstacles},
  author={Muguira-Iturralde, Jose and Curtis, Aidan and Du, Yilun and Kaelbling, Leslie Pack and Lozano-P{\'e}rez, Tom{\'a}s},
  booktitle={2023 IEEE International Conference on Robotics and Automation (ICRA)},
  pages={10083--10089},
  year={2023},
  organization={IEEE}
}

@inproceedings{zhou2024llm,
  title={Llm-bt: Performing robotic adaptive tasks based on large language models and behavior trees},
  author={Zhou, Haotian and Lin, Yunhan and Yan, Longwu and Zhu, Jihong and Min, Huasong},
  booktitle={2024 IEEE International Conference on Robotics and Automation (ICRA)},
  pages={16655--16661},
  year={2024},
  organization={IEEE}
}
}


\appendix

\subsection{Prompts}
\label{appendix: prompts}

\subsubsection{Object-Level Proposer}
You are a quadruped robot on the ground in a 3D world. Your goal is to navigate to a specific point in the 3D space. Your navigation goal is \textit{\textcolor{blue}{[goal point]}}, and your current scene understanding is \textit{\textcolor{blue}{[scene understanding]}}.
There are several objects in the scene that you may utilize. We use two parameters, position and size, to represent the location and size of an object, respectively. 
Each object's position is represented by a 3D vector [x, y, z]. Each object's size is represented as a 3D vector [length, width, height]. \textit{\textcolor{blue}{[object description]}}.
You have a skill library containing the following skills and corresponding parameters: 
\textit{\textcolor{blue}{[skill library]}}.
You must follow these constraints:
\textit{\textcolor{blue}{[constraints]}}.
Give me five different abstract plans for using objects to help you complete navigation tasks. The plan must include which objects you need to use, the sequence you use the objects, how to use the objects. You must analyze the problem step by step and show the thinking process.
You must follow the following answer template: 

[begin of plan]

Plan1: I need to use [objects]. First, ... Second ...

Plan2: ...

[end of plan]

\subsubsection{Skill-Level Proposer}
You are a quadruped robot on the ground. Your goal is to navigate to a specific point in the 3D space. Your navigation goal is \textit{\textcolor{blue}{[goal point]}}.
You have a skill library containing the following skills and corresponding parameters: 
\textit{\textcolor{blue}{[skill library]}}.
Here are some abstract plans: \textit{\textcolor{blue}{[object-level plans]}}, you need to generate detailed plans according to each abstract plan. Each step in the detailed plan consists of a skill with an abstract position like 'walk-to('abstract position')'. 

You must follow the following answer template: 

[begin of plan]

Plan1: [

('step1','<skill>'),

...

],

[end of plan]

\subsubsection{Evaluator}
You are a quadruped robot on the ground in a 3D world. Your goal is to navigate to a specific point in the 3D space. Your navigation goal is \textit{\textcolor{blue}{[goal point]}}, and your current scene understanding is \textit{\textcolor{blue}{[scene understanding]}} 
There are several objects in the scene that you may utilize: \textit{\textcolor{blue}{[object description]}}.
You have a skill library containing the following skills and corresponding parameters: \textit{\textcolor{blue}{[skill library]}}. 
Here are some detailed plans to accomplish the navigation task: ...
You need to evaluate the reward of each step in the plan based on its contribution to task completion, assigning a value between 0 and 1. Specifically, this involves simulating whether future steps can reach the goal under the current step to assess its impact. If the step does not satisfy the given constraints, the reward is 0.
Constraints you must follow: \textit{\textcolor{blue}{[constraints]}}.
At the end of your response, reply to me with the following answer template:

[begin of evaluation]

Plan1: [

('step1', '<skill>', 'reward: ...'),

...

],

[end of evaluation]

\subsubsection{Parameter Calculation}
The current plan is..., you need to calculate the 3D coordinates [x, y, z] of the abstract position in the plan based on the robot pose and the updated object information.
The robot pose is \textit{\textcolor{blue}{[robot pose]}}.
The object information is \textit{\textcolor{blue}{[object description]}}.
You need to consider the spatial relationship between objects to obtain a reasonable position.
You must calculate the position along each dimension step by step.

\subsubsection{Advisor}
You are a quadruped robot on the ground in a 3D world. Your goal is to navigate to a specific point in the 3D space. Your navigation goal is \textit{\textcolor{blue}{[goal point]}}. There are several objects in the scene that you may utilize: \textit{\textcolor{blue}{[object description]}}.
Your current plan is: ... 
Now, you have a new observation interpretation of the environment: ...
You need to determine whether to replan to modify your current plan based on the current plan, this new environmental observation, and the following criteria. You must analyze whether to reply, provide me with the reason, and respond with "Yes" or "No".

Replanning criteria:

* If the new observation is an execution failure of the current plan, then replanning is necessary. 

* If the new observation is a new object, you need to evaluate how this new object might help complete your task. If using it results in a more effective plan than your current one, you need to replan.

* If the new observation is a revaluation of a previously known object, you need to determine whether this new information impacts your current plan. For example, if your plan requires climbing onto a box, but new observations show that the box is too high to climb, you need to replan.

\subsubsection{Arborist}
You are a quadruped robot on the ground in a 3D world. Your goal is to navigate to a specific point in the 3D space. Your navigation goal is \textit{\textcolor{blue}{[goal point]}}. There are several objects in the scene that you may utilize: \textit{\textcolor{blue}{[object description]}}. Constraints you must follow: \textit{\textcolor{blue}{[constraints]}}. Your current plan is: ... Now, you have a new observation of the environment: .... Here are replanning suggestions by the advisor: ...
You must make adjustments to your current plans based on the suggestions provided by the advisor, such as expanding new skills for new objects or pruning infeasible skills.


\subsection{Reward Design}
\label{appendix: reward-design}

We list the symbol definition and the reward terms of the four skills in the pre-trained skill library as follows, with $\phi(x):=\exp(-\frac{||x||^2}{0.25})$.

\begin{table}[H]\centering
\footnotesize
\caption{Definition of symbols.}
\begin{tabular}{rl}
    \toprule
    \textbf{Symbol} & \textbf{Description}\\
     \midrule
     $\mathbf{q}_j$ & Joint positions \\
     $\dot{\mathbf{q}_j}$ & Joint velocities \\
     $\ddot{\mathbf{q}_j}$ & Joint accelerations \\
     ${\mathbf{q}^*_j}$ & Target joint positions \\
     ${\mathbf{q}^{*,pre}_j}$ & Last target joint positions\\
     $\boldsymbol{\tau}_j$ & Joint torques\\
     $\mathbf{v}_{b}$ & Base linear velocity\\
     $\boldsymbol{\omega}_{b}$ & Base angular velocity \\
     $\boldsymbol{x}_{b}$ & Base position\\
     $\boldsymbol{\theta}_{b}$ & Base heading\\
     $\boldsymbol{x}_{o}$ & Object pose\\
     $\mathbf{v}^*_{b}$ & Commanded base linear velocity \\
     $\boldsymbol{\omega}^*_{b}$ & Commanded base angular velocity \\
     $\boldsymbol{x}^*_{b}$ & Commanded base position\\
     $\boldsymbol{x}^*_{o}$ & Commanded object pose\\
     $n_{c}$ & Number of collisions \\
     $\mathbf{t}_{air, f}$ & Feet air time \\
 \bottomrule
\end{tabular}
\label{table:nomenclature}
\end{table}


\begin{table}[H]\centering
\footnotesize
\caption{Reward terms in walking skill.}
\begin{tabular}{rcl}
    \toprule
     \textbf{Reward terms} & \textbf{Definition} & \textbf{Weight}\\
     \midrule
     Linear velocity tracking & $\phi(\mathbf{v}^*_{b,xy} - \mathbf{v}_{b,xy})$ & $1 $ \\
     Angular velocity tracking & $\phi(\boldsymbol{\omega}^*_{b,z} - \boldsymbol{\omega}_{b,z})$ & $0.5 $\\
     Linear velocity penalty & $-\mathbf{v}_{b,z}^2$ & $2 $ \\
     Angular velocity penalty & $-||\boldsymbol{\omega}_{b,xy}||^2$ & $0.05 $\\
     Joint torques & $-||\boldsymbol{\tau}_j||^2$ & $0.00001 $\\
     Joint accelerations & $-||\ddot{\mathbf{q}_j}||^2$ & $2.5*10^{-7} $ \\
     Action rate & $-||\mathbf{q}^*_j-\mathbf{q}^{*,pre}_j||^2$ & $0.01 $ \\
     Collisions & $-n_{collision}$ & $1 $\\
     Feet air time & $ \sum_{f=0}^{4}(\mathbf{t}_{air, f} - 0.5)$ & $0.125 $\\
 \bottomrule\\
\end{tabular}
\label{table:rewards_walk}
\end{table}

\begin{table}[H]\centering
\footnotesize
\caption{Reward terms in climbing skill.}
\begin{tabular}{rcl}
    \toprule
      \textbf{Reward terms} & \textbf{Definition} & \textbf{Weight}\\
     \midrule
      Position tracking & $1-0.5 ||\boldsymbol{x}^*_{b} - \boldsymbol{x}_{b}||$ & $1 $ \\
      Move direction & $\text{cos}\langle\mathbf{v}_{b}, \boldsymbol{x}^*_{b} - \boldsymbol{x}_{b}\rangle$ & $2 $ \\
     Linear velocity penalty & $-\mathbf{v}_{b,z}^2$ & $2 $ \\
     Angular velocity penalty & $-||\boldsymbol{\omega}_{b,xy}||^2$ & $0.05 $\\
     Joint torques & $-||\boldsymbol{\tau}_j||^2$ & $0.00001 $\\
     Joint accelerations & $-||\ddot{\mathbf{q}_j}||^2$ & $2.5*10^{-7} $ \\
     Action rate & $-||\mathbf{q}^*_j-\mathbf{q}^{*,pre}_j||^2$ & $0.01 $ \\
     Collisions & $-n_{collision}$ & $1 $\\
     Feet air time & $ \sum_{f=0}^{4}(\mathbf{t}_{air, f} - 0.5)$ & $0.125 $\\
 \bottomrule\\
\end{tabular}
\label{table:rewards_climb}
\end{table}

\begin{table}[H]\centering
\footnotesize
\caption{Reward terms in navigation skill.}
\begin{tabular}{ccc}
    \toprule
      \textbf{Reward terms} & \textbf{Definition} & \textbf{Weight}\\
     \midrule
      Position tracking & $1-0.5 ||\boldsymbol{x}^*_{b} - \boldsymbol{x}_{b}||$ & $5 $ \\
      Negative x-velocity penalty & $\max(\mathbf{v}_{b,x},0)$ & $2 $ \\
      Collisions & $-n_{collision}$ & $10 $\\
 \bottomrule\\
\end{tabular}
\label{table:rewards_navigation}
\end{table}

\begin{table}[H]\centering
\footnotesize
\caption{Reward terms in pushing skill.}
\begin{tabular}{ccc}
    \toprule
      \textbf{Reward terms} & \textbf{Definition} & \textbf{Weight}\\
     \midrule
      Object Position tracking & $1-||\boldsymbol{x}^*_{o, xy} - \boldsymbol{x}_{o, xy}||$ & $2 $ \\
      Object heading tracking & $1-||\boldsymbol{x}^*_{o, h} - \boldsymbol{x}_{o, h}||$ & $1 $ \\
      Negative x-velocity penalty & $\max(\mathbf{v}_{b,x},0)$ & $1 $ \\
      Face to object & $\text{cos}\langle \mathbf{\theta}_{b}, \boldsymbol{x}_{o} - \boldsymbol{x}_{b}\rangle$ & $2 $ \\
 \bottomrule\\
\end{tabular}
\label{table:rewards_pushing}
\end{table}

\subsection{Performance of Primitive Skills}

\begin{figure}[!ht]
    \centering
    \begin{subfigure}[b]{0.37\textwidth}
        \includegraphics[width=\textwidth]{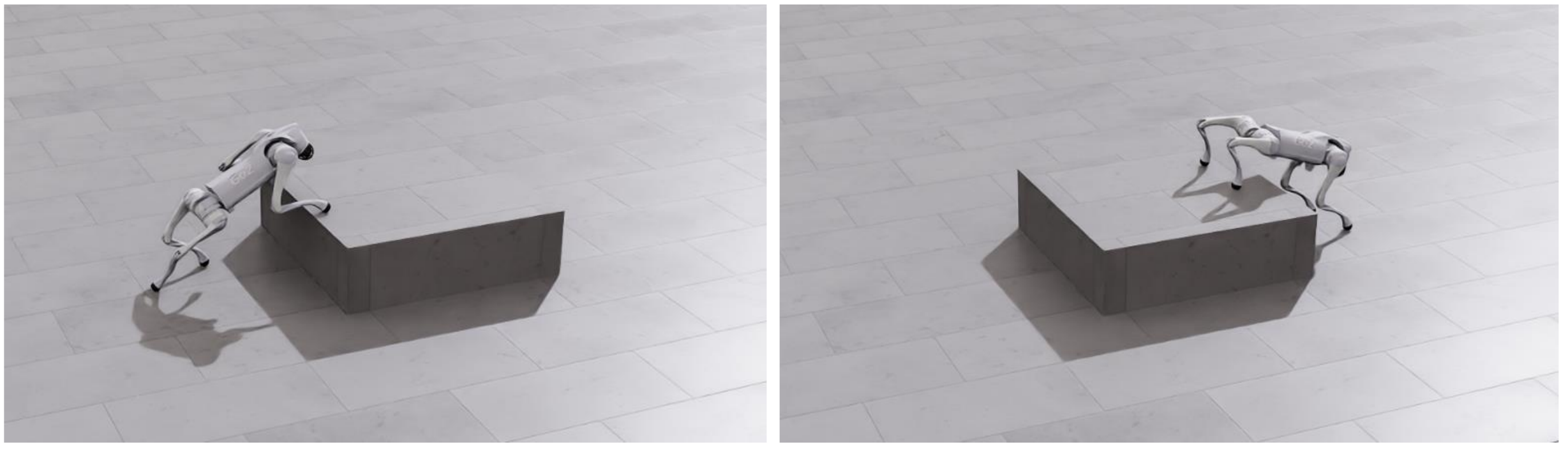}
        \caption{Climbing up/down skill}
        \label{fig:climb_skill}
    \end{subfigure}
    \hfill
    \begin{subfigure}[b]{0.37\textwidth}
        \includegraphics[width=\textwidth]{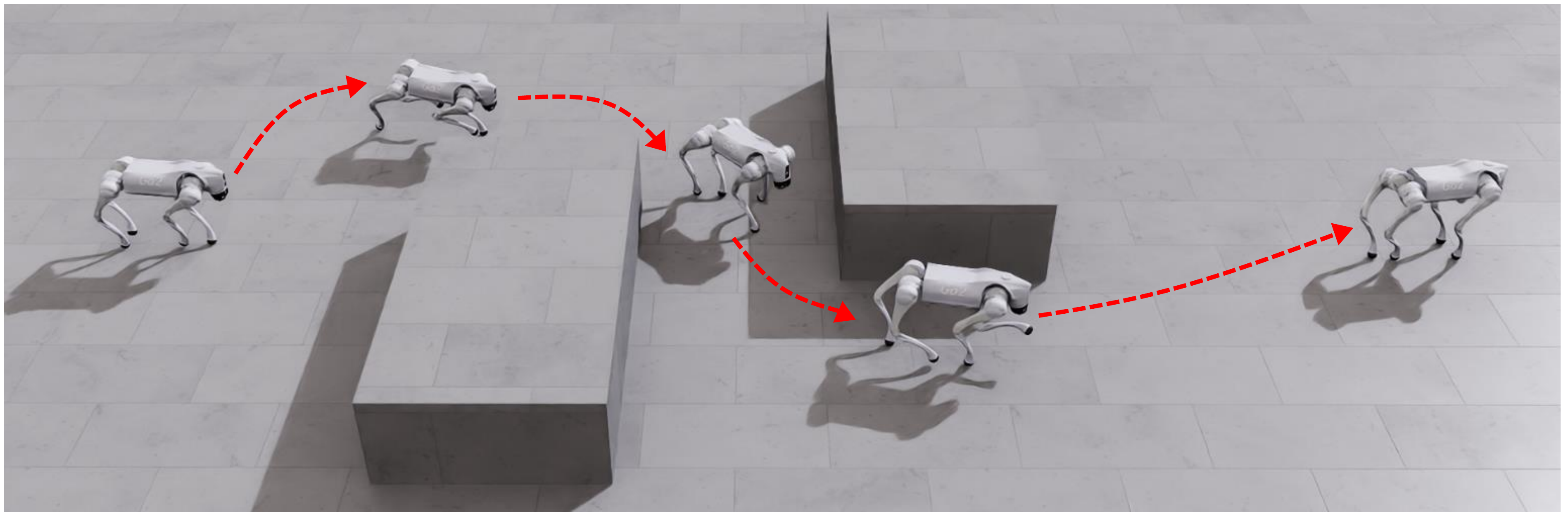}
        \caption{Navigation skill}
        \label{fig:navigation_skill}
    \end{subfigure}
    

    \begin{subfigure}[b]{0.37\textwidth}
        \includegraphics[width=\textwidth]{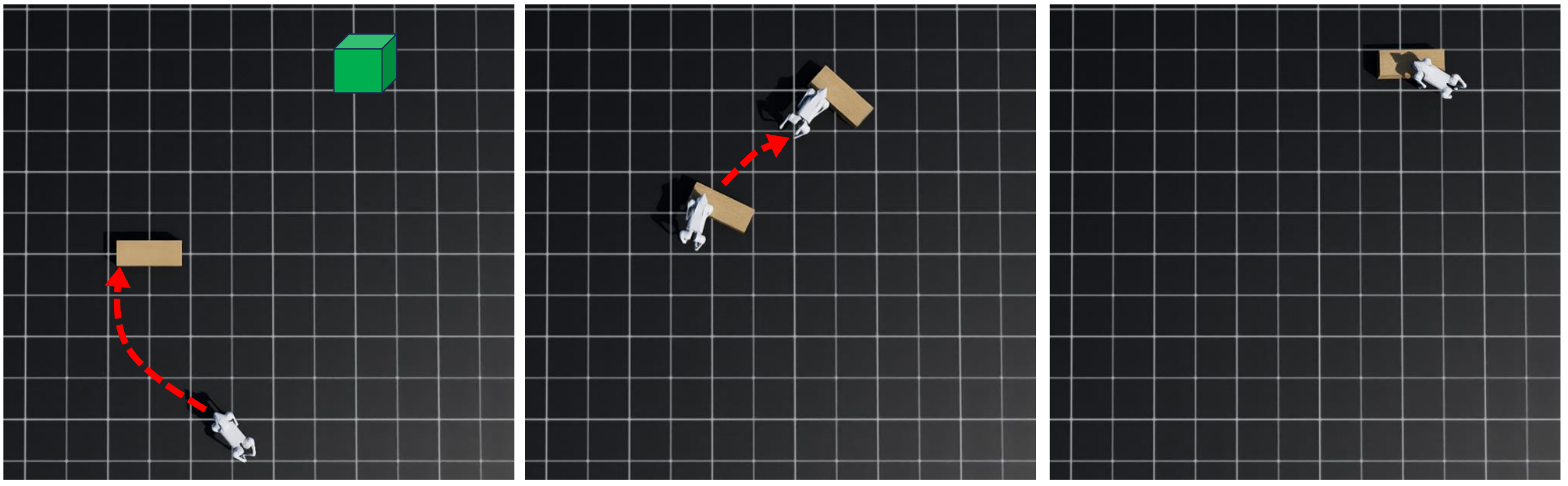}
        \caption{Pushing skill: small rectangle box}
        \label{fig:push_skill1}
    \end{subfigure}
    \hfill
    \begin{subfigure}[b]{0.37\textwidth}
        \includegraphics[width=\textwidth]{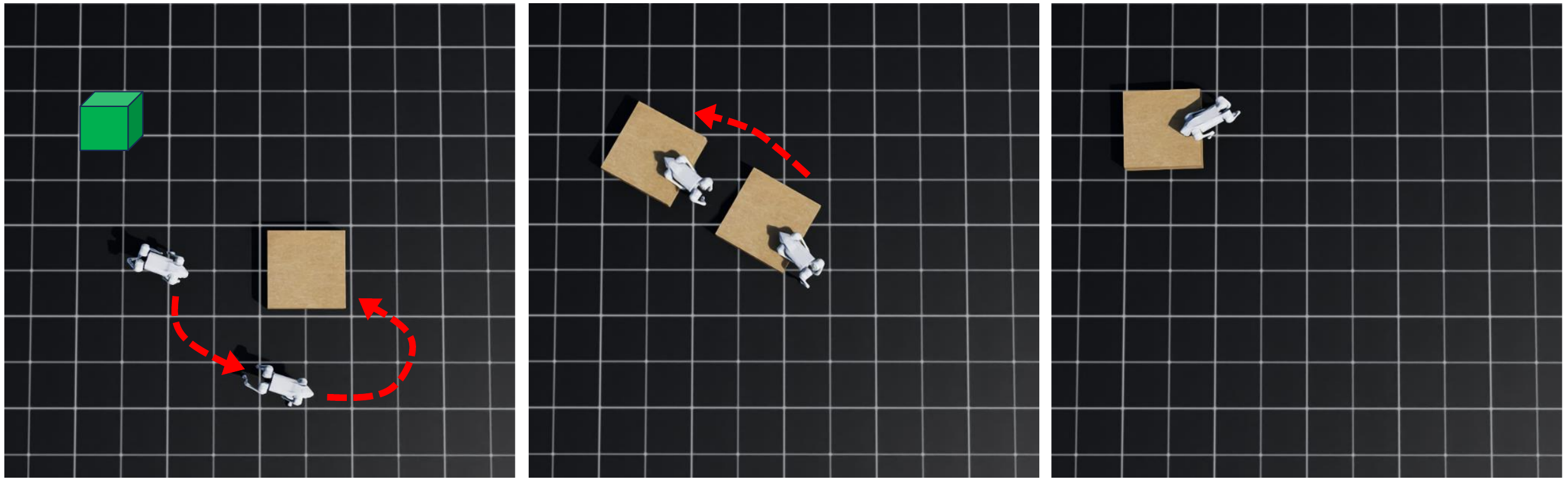}
        \caption{Pushing skill: big square box}
        \label{fig:push_skill2}
    \end{subfigure}
    
    \caption{Display of diverse RL-driven skills.
    (a) and (b) show the climbing and navigation skills adapted to various terrains, while (c) and (d) showcase robust interaction capabilities with boxes of different shapes and sizes.
    }
    \label{fig:skill_lib}
\end{figure}

We visualize the performance of various skills in \Cref{fig:skill_lib}.
In \Cref{fig:climb_skill}, the robot is depicted climbing up and down a 0.3-meter height, effectively adapting to different terrains to accomplish navigation tasks.
\Cref{fig:navigation_skill} illustrates the robot's capability to navigate to a target point while avoiding obstacles in a cluttered environment.
In \Cref{fig:push_skill1} and \Cref{fig:push_skill2}, we showcase the robot pushing boxes of varying sizes and shapes to the designated positions, demonstrating the generalization capability of the pushing strategy. 
These versatile skills collectively form a comprehensive skill library that enables the robot to handle various sub-tasks in complex navigation scenarios.

\subsection{Case Visualization with Concrete Prompts}
\label{appendix: case-visualization}

We provide a detailed process for the obstacle traversal scenario in real-world experiments, including prompts and outputs for all LLM-based modules. 
This covers the initial task decomposition and the replanning stage in case of failure. 
Due to space limitations, we only present the most informative content in prompt and LLM output.

\begin{figure}[!h] 
\centering
\includegraphics[width=0.85\linewidth]{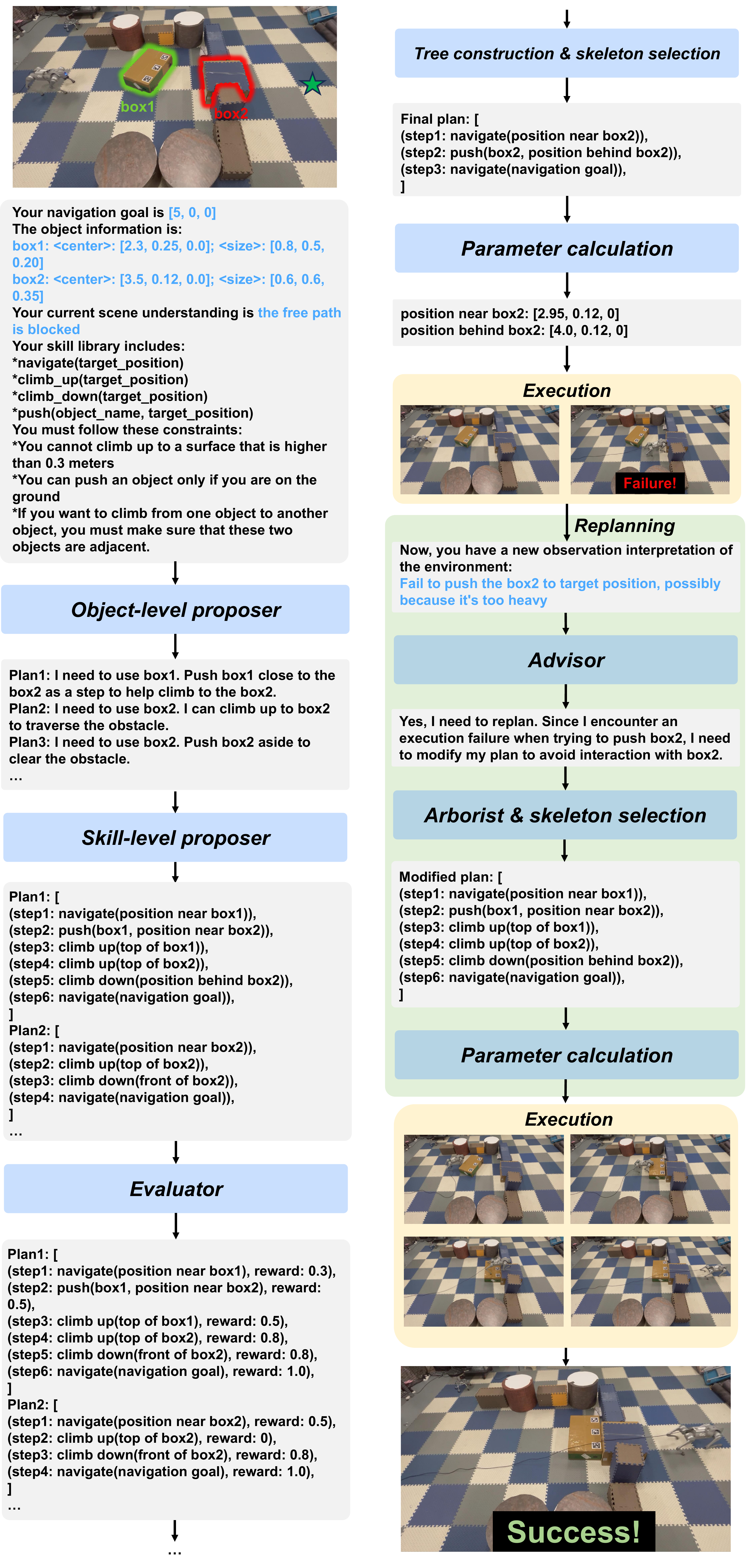} 
\caption{Visualization of prompts and output of each module.}
\label{Fig: case_visualization}
\end{figure}

\end{document}